\documentclass[manuscript]{acmart}

\AtBeginDocument{%
  \providecommand\BibTeX{{%
    \normalfont B\kern-0.5em{\scshape i\kern-0.25em b}\kern-0.8em\TeX}}}

\setcopyright{acmcopyright}
\copyrightyear{2022}
\acmYear{2022}
\acmJournal{TIST}
\acmMonth{}
\acmVolume{}
\acmNumber{}
\acmArticle{}
\acmArticleSeq{}
\acmSubmissionID{}

\acmPrice{}
\acmISBN{}

\usepackage{amsmath}
\newcommand{\R}{\mathbb{R}}
\newtheorem{definition}{Definition}
\usepackage{multirow}
\begin{document}

\title{A Survey on Graph Representation Learning Methods}


\author{Shima Khoshraftar}
\email{khoshraf@eecs.yorku.ca}
\author{Aijun An}
\email{aan@eecs.yorku.ca}
\affiliation{%
  \institution{Electrical Engineering and Computer Science Department, York University}
  \streetaddress{Keele Street}
  \city{Toronto}
  \country{Canada}}


\renewcommand{\shortauthors}{S. Khoshraftar and A. An.}

\begin{abstract}
  Graphs representation learning has been a very active research area in recent years. The goal of graph representation learning is to generate graph representation vectors that capture the structure and features of large graphs accurately. This is especially important because the quality of the graph representation vectors will affect the performance of these vectors in downstream tasks such as node classification, link prediction and anomaly detection.
  Many techniques have been proposed for generating effective graph representation vectors, which generally fall into two categories: 
  traditional graph embedding methods and graph neural nets (GNN) based methods. 
  These methods can be applied to both static and dynamic graphs. A static graph is a single fixed graph, while a dynamic graph evolves over time and its nodes and edges can be added or deleted from the graph. In this survey, we review
  the graph embedding methods in both traditional and GNN-based categories for both static and dynamic graphs and include the recent papers published until the time of submission. 
  In addition, we summarize a number of limitations of GNNs and the proposed solutions to these limitations. Such a summary has not been provided in previous surveys. Finally, we explore some open and ongoing research directions for future work.
  
  \end{abstract}


\maketitle

\section{Introduction}
\label{sec:intro}
 Graphs are powerful data structures to represent networks that contain entities and relationships between entities. There are very large networks in different domains including social networks, financial transactions and biological networks. For instance, in social networks people are the nodes and their friendships constitute the edges. In financial transactions, the nodes and edges could be people and their money transactions. One of the strengths of a graph data structure is its generality, meaning that the same structure can be used to represent different networks. In addition, graphs have strong foundations in mathematics, which could be leveraged for analyzing and learning from complex networks. 

In order to use graphs in different downstream applications, it is important to represent them effectively. Graph can be simply represented using the adjacency matrix which is a square matrix whose elements indicate whether pairs of vertices are adjacent or not in the graph, or using the extracted features of the graph. However, the dimensionality of the adjacency matrix is often very high for big graphs, and the feature extraction based methods are time consuming and may not represent all the necessary information in the graphs. Recently, the abundance of data and computation resources paves the way for more flexible graph representation methods. Specifically, graph embedding methods have been very successful in graph representation. These methods project the graph elements (such as nodes, edges and subgraphs) to a lower dimensional space and preserve the properties of graphs.  
Graph embedding methods can be categorized into traditional graph embedding and graph neural net (GNN) based graph embedding methods. Traditional graph embedding methods capture the information in a graph by applying different techniques including random walks, factorization methods and non-GNN based deep learning. These methods can be applied to both static and dynamic graphs. A static graph is a single fixed graph, while a dynamic graph evolves over time and its nodes and edges can be added or deleted from the graph. For example, a molecule can be represented as a static graph, while a social network can be represented by a dynamic graph. 
Graph neural nets are another category of graph embedding methods that have been proposed recently. In GNNs, node embeddings are obtained by aggregating the embeddings of the node's neighbors. Early works on GNNs were based on recurrent neural networks. However, later convolutional graph neural nets were developed that are based on the convolution operation. In addition, there are spatial-temporal GNNs and dynamic GNNs that leverage the strengths of GNNs in evolving networks. 


In this survey, we conduct a review of both traditional and GNN-based graph embedding methods in static and dynamic settings. To the best of our knowledge, there are several other surveys and books on graph representation learning. The surveys in \cite{hamilton2017representation,goyal2018graph,cai2018comprehensive,HamiltonBook,chen2020graph,zhang2020deep,zhou2020graph,ma2021deep,xia2021graph,zhang2018network,amara2021network,liu2021network,xu2021understanding,li2020network,zhou2022network} mainly cover the graph embedding methods for static graphs and their coverage of dynamic graph embedding methods is limited if any. Inversely, \cite{kazemi2020representation,xie2020survey,skardinga2021foundations,barros2021survey} surveys mainly focus on dynamic graph embedding methods. \cite{georgousis2021graph,wu2020comprehensive} review both the static and dynamic graph embedding methods but they focus on GNN-based methods. The distinction between our survey and others are as follows:
\begin{enumerate}
    \item We put together the graph embedding methods in both traditional and GNN-based categories for both static and dynamic graphs and include 
    over 300 papers consisting of papers 
    published in reputable venues in data mining, machine learning and artificial intelligence \footnote{The venues include KDD, ICLR, ICML, NeurIPS, AAAI, IJCAI, ICDM, WWW, WSDM, DSAA, SDM, CIKM.} since 2017 until the time of this submission, and also influential papers with high citations published before 2017. 
    
    \item We summarize a number of limitations of GNN-based methods and the proposed solutions to these limitations until the time of submission. These limitations are expressive power, over-smoothing, scalability, over-squashing, capturing long-range dependencies, design space, neglecting substructures, homophily assumptions, and catastrophic forgetting. Such a summary was not provided in previous surveys.
    \item We provide a list of the real-world applications of GNN-based methods that are deployed in production.
    \item We suggest a list of future research directions including new ones that are not covered by previous surveys. 
\end{enumerate}

The organization of the survey is as follows. First, we provide basic background on graphs. Then, the traditional node embedding methods for static and dynamic graphs are reviewed. After that, we survey static, spatial-temporal and dynamic GNN-based graph embedding methods and their real-world applications. Then, we continue by summarizing the limitations of GNNs. Finally, we discuss the ongoing and future research directions in the graph representation learning area.

\section{Graphs}
\label{sec:first}
Graphs are powerful tools for representing entities and relationships between them. Graphs have applications in many domains including social networks, E-commerce and citation networks. In social networks such as Facebook, nodes in the graph are the people and the edges represent the friendship between them. In E-commerce, the Amazon network is a good example, in which users and items are the nodes and the buying or selling relationships are the edges. 


\begin{definition}
Formally, a graph $G$ is defined as a tuple $G=(V,E)$ where $V = \{v_0, v_1, ..., v_n\}$ is the set of $n$ nodes/vertices and $E = \{e_0, e_1,..., e_m \} \subseteq V \times V$ is the set of $m$ edges/links of $G$, where an edge connects two vertices.
\end{definition}
A graph can be {\it directed} or {\it undirected}. In a directed graph, an edge $e_k=(v_i, v_j)$ has a direction with $v_i$ being the starting vertex and $v_j$ the ending vertex.
Graphs can be represented by their adjacency, degree and Laplacian matrices, which are defined as follows:

\begin{definition}
The {\it adjacency matrix} $A$ of a graph $G$ with $n$ vertices is an $n\times n$ matrix, where an element $a_{ij}$ in the matrix equals to 1 if there is an edge between node pair $v_i$ and $v_j$ and is 0 otherwise. An adjacency matrix can be {\it weighted} in which the value of an element represents the weight (such as importance) of the edge it represents.
\end{definition}
\begin{definition}
The {\it degree matrix} $D$ of a graph $G$ with $n$ vertices is an $n\times n$ diagonal matrix, where an element $d_{ii}$ is the degree of node $v_i$ for $i=\{1,...,n\}$ and all other $d_{ij}=0$. In undirected graphs, where edges have no direction, the degree of a node refers to the number of edges attached to that node. For directed graphs, the degree of a node can be the number of incoming or outgoing edges of that node, resulting in an {\it in-degree} or {\it out-degree} matrix, respectively.
\end{definition}
\begin{definition}  
The Laplacian matrix $L$ of a graph $G$ with $n$ vertices is an $n\times n$ matrix, defined as $L = D - A$, where $D$ and $A$ are $G$'s degree and adjacency matrix, respectively. 
\end{definition}

\subsection{Graph Embedding}
In order to use graphs in downstream machine learning and data mining applications, graphs and their entities such as nodes and edges need to be represented using numerical features. 
One way to represent a graph is its adjacency matrix. However, an adjacency matrix is memory-consuming for representing very large graphs because its size is $|V| \times |V|$.
We can represent a graph and its elements using their features. 
Especially, a node in the graph can be represented with a set of features that could help the performance of the representation in a particular application. For example, in anomaly detection application, the nodes with the densest neighborhood have the potential to be anomalous. Therefore, if we include the in-degree and out-degree of nodes in the node representation, we can more likely detect the anomalous nodes with high accuracy because the anomalous nodes often have larger degrees.  However, it could be hard to find features that are important in different applications and can also represent the entire structure of the graph. In addition, it is time consuming to extract these features manually.
Therefore, the graph embedding methods have been proposed, which study the issue of automatically generating representation vectors for the graphs. 
These methods formulate the graph representation learning as a machine learning task and generate embedding vectors leveraging the structure and properties of the graph as input data. Graph embedding techniques include node, edge and subgraph embedding techniques, which are defined as follows.

\begin{definition}
(Node embedding). Let $G=(V,E)$ be a graph, where $V$ and $E$ are the set of nodes and the set of edges of the graph, respectively. Node embedding learns a mapping function $f:v_i \rightarrow \R^d$ that encodes each graph's node $v_i$ into a low dimensional vector of dimension $d$ such that $d<<|V|$ and the similarities between nodes in the graph are preserved in the embedding space.
\end{definition}

\begin{figure}
    \centering
    \includegraphics[scale = .6]{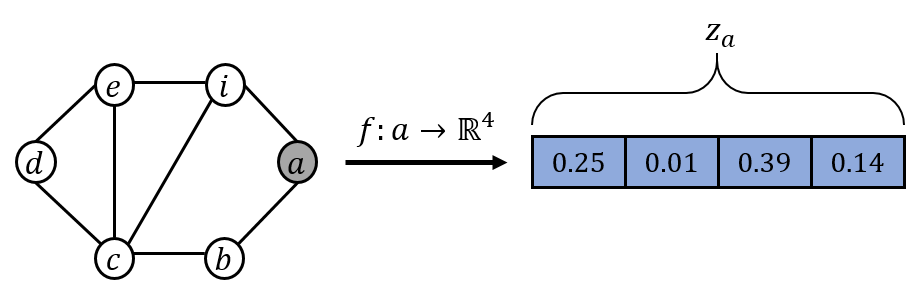}
    \caption{The graph on the left hand side consists of 6 nodes $\{a,b,c,d,e,i\}$ and 8 edges. Graph embedding methods map each node of the graph into an embedding vector with dimension $d$. For the demonstration purpose, the node $a$ is embedded into an embedding vector $z_a$ of dimension 4 with given values.}
    \label{fig7}
\end{figure}

Figure \ref{fig7} shows a sample graph and that an embedding method maps node $a$ in the graph to a vector of dimension 4. 

\begin{definition}
(Edge embedding). Let $G=(V,E)$ be a graph, where $V$ and $E$ are the set of nodes and the set of edges of the graph, respectively. Edge embedding converts each edge of $G$ into a low dimensional vector of dimension $d$ such that $d<<|V|$ and the similarities between edges in the graph are preserved in the embedding space.
\end{definition}
While edge embeddings can be learned directly from graphs, most commonly they are derived from node embeddings. For example, let $(v_i,v_j) \in E$ be an edge between two nodes $v_i$ and $v_j$ in a graph $G$ and $z_i,z_j$ be the embedding vectors for nodes $v_i, v_j$. An embedding vector for the edge $(v_i,v_j)$ can be obtained by applying a binary operation such as hadamard product, mean, weighted-L1 and weighted-L2 on the two node embedding vectors $z_i$ and $z_j$ \cite{grover2016node2vec}. 

\begin{definition}
(Subgraph embedding). Let $G=(V,E)$ be a graph. Subgraph embedding techniques in machine learning convert a subgraph of $G$ into a low dimensional vector of dimension $d$ such that $d<<|V|$ and the similarities between subgraphs are preserved in the embedding space.
\end{definition}
A subgraph embedding vector is usually created by aggregating the embeddings of the nodes in the subgraph using aggregators such as a mean operator. 

As node embeddings are the building block for edge and subgraph embeddings, almost all the graph embedding techniques developed so far are node embedding techniques. Thus, the embedding techniques we describe in this survey are mostly node embedding techniques unless otherwise stated.

\subsection{Graph Embedding Applications}
The generated embedding vectors can be utilized in different applications including node classification, link prediction and graph classification. Here, we explain some of these applications.


\textbf{Node Classification.}
Node classification task assigns a label to the nodes in the test dataset. This task has many applications in different domains. For instance, in social networks, a person's political affiliation can be predicted based on his friends in the network. In node classification, each instance in the training dataset is the node embedding vector and the label of the instance is the node label. Different regular classification methods such as Logistic Regression and Random Forests can be trained on the training dataset and generate the node classification scores for the test data. Similarly, Graph classification can be performed using graph embedding vectors.

\textbf{Link Prediction.} 
Link prediction is one of the important applications of node embedding methods. It predicts the likelihood of an edge formation between two nodes.
Examples of this task include recommending friends in social networks and finding biological connections in biological networks. Link prediction can be formulated as a classification task that assigns a label for edges. Edge label 1 means that an edge is likely to be created between two nodes and the label is 0 otherwise. For the training step, a sample training set is generated using positive and negative samples. Positive samples are the edges the exist in the graph. Negative samples are the edges that do not exist and their representation vector can be generated using the node vectors. Similar to node classification, any classification method can be trained on the training set and predict the edge label for test edge instances.

\textbf{Anomaly Detection.} 
Anomaly detection is another application of node embedding methods. The goal of anomaly detection is to detect the nodes, edges, or graphs that are anomalous and the time that anomaly occurs. Anomalous nodes or graphs deviate from normal behavior. For instance, in banks' transaction networks, people who suddenly send or receive large amounts of money or create lots of connections with other people could be potential anomalous nodes.  An anomaly detection task can be formulated as a classification task such that each instance in the dataset is the node representation and the instance label is 0 if the node is normal and 1 if the node is anomalous. This formulation needs that we have a dataset with true node labels. One of the issues in anomaly detection is the lack of datasets with true labels. An alleviation to this issue in the literature is generating synthetic datasets that model the behaviors of real world datasets. Another way to formulate the anomaly detection problem, especially in dynamic graphs, is viewing the problem as a change detection task. In order to detect the changes in the graph, one way is to compute the distance between the graph representation vectors at consecutive times. The time points that the value of this difference is far from the previous normal values, a potential anomaly has occurred \cite{goyal2017dyngem}.

\textbf{Graph Clustering.} 
In addition to classification tasks, graph embeddings can be used in clustering tasks as well. This task can be useful in domains such as social networks for detecting communities and biological networks to identify similar groups of proteins. Groups of similar graphs/node/edges can be detected by applying clustering methods such as the Kmeans method \cite{macqueen1967some} on the graph/node/edge embedding vectors.


\textbf{Visualization.} 
One of the applications of node embedding methods is graph visualization because node embedding methods map nodes in lower dimensions and the nodes, edges, communities and different properties of graphs can be better seen in the embedding space. Therefore, graph visualization is very helpful for the research community to gain insight into graph data, especially very large graphs that are hard to visualize.  




\begin{figure}
    \centering
    \includegraphics[scale=0.6]{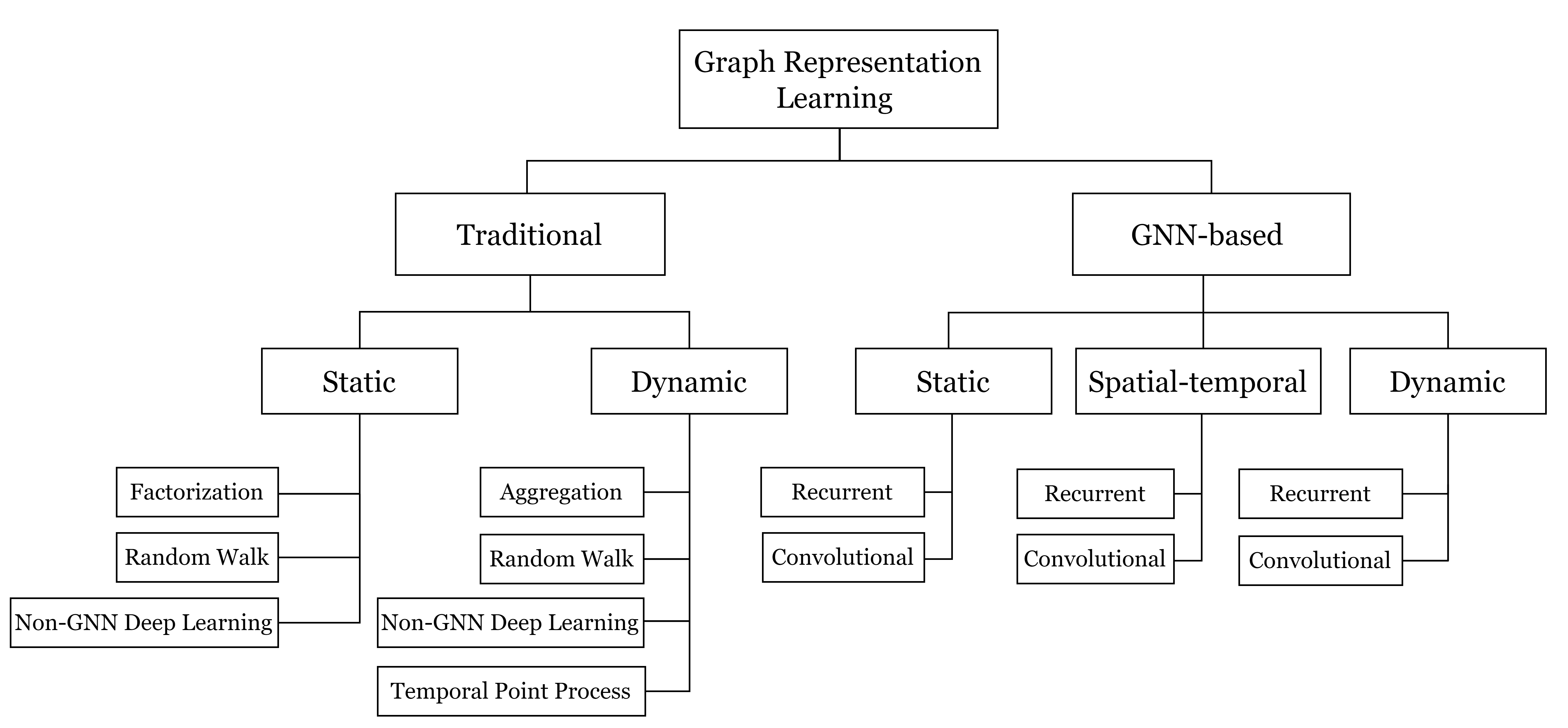}
    \caption{Categories of Graph Representation Learning Methods.}
    \label{cat}
\end{figure}

\begin{table}[]
\centering
\caption{Graph embedding methods in both traditional and GNN-based categories. Trad and GNN stand for Traditional and GNN-based graph embedding.}
\label{table:methods}
\begin{tabular}{l|p{1.3cm}|p{12.2cm}}
\hline
Type & \multicolumn{1}{c|}{Graph} & \multicolumn{1}{c}{Methods} \\ \hline
\multirow{2}{*}{Trad} & Static &  Node2vec
\cite{grover2016node2vec}, Deepwalk \cite{perozzi2014deepwalk}, Graph Factorization \cite{ahmed2013distributed}, GraRep \cite{cao2015grarep}, HOPE \cite{ou2016asymmetric}, STRAP \cite{yin2019scalable}, HARP \cite{chen2018harp}, LINE \cite{tang2015line}, SDNE \cite{wang2016structural}, DNGR \cite{cao2016deep}, VGAE \cite{kipf2016variational},
AWE \cite{ivanov2018anonymous}, PRUNE \cite{lai2017prune}, E[D] \cite{abu2018watch}, ULGE \cite{nie2017unsupervised}, APP \cite{zhou2017scalable}, CDE \cite{li2018community}, GNE \cite{du2018galaxy}, DNE \cite{shen2018discrete}, DANE \cite{gao2018deep}, RandNE \cite{zhang2018billion}, SANE \cite{wang2018united}, BANE \cite{yang2018binarized}, LANE \cite{huang2017label}, VERSE \cite{tsitsulin2018verse}, ANECP \cite{huang2020attributed}, NOBE \cite{jiang2018spectral}, AANE \cite{huang2017accelerated}, Reinforce2vec \cite{xiao2020vertex}, REFINE \cite{zhu2021refine}, M-NMF \cite{wang2017community}, struct2vec \cite{ribeiro2017struct}, SNEQ \cite{he2020sneq}, PAWINE \cite{wang2020bringing}, FastRP \cite{chen2019fast}, SNS \cite{lyu2017enhancing}, InfiniteWalk \cite{chanpuriya2020infinitewalk}, EFD \cite{chanpuriya2020node}, NetMF \cite{qiu2018network}, Lemane \cite{zhang2021learning}, AROPE \cite{zhang2018arbitrary}, NetSMF \cite{qiu2019netsmf}, SPLITTER \cite{epasto2019single}, Ddgk \cite{al2019ddgk}, GVNR \cite{brochier2019global}, LouvainNE \cite{bhowmick2020louvainne}, HONE \cite{rossi2020structural}, CAN \cite{meng2019co},
Methods in \cite{yang2017fast,huang2021broader,liu2019general}\\ \cline{2-3} 
 & Dynamic & CTDNE \cite{nguyen2018continuous}, DynNode2vec \cite{mahdavi2018dynnode2vec}, LSTM-Node2vec \cite{khoshraftar2019dynamic}, EvoNRL \cite{heidari2020evolving}, DynGEM \cite{goyal2017dyngem}, Dyn-VGAE \cite{mahdavi2019dynamic}, DynGraph2vec \cite{goyal2020dyngraph2vec}, HTNE \cite{zuo2018embedding}, DynamicTriad \cite{zhou2018dynamic}, DyRep \cite{trivedi2019dyrep}, MTNE \cite{huang20motif}, DNE \cite{du2018dynamic}, Toffee \cite{ma2021temporal}, HNIP \cite{qiu2020temporal}, tdGraphEmbed \cite{beladev2020tdgraphembed}, DRLAN \cite{liu2020dynamic}, TIMERS \cite{zhang2018timers}, M2DNE \cite{lu2019temporal}, DANE \cite{li2017attributed}, TVRC \cite{sharan2008temporal}, tNodeEmbed \cite{singer2019node}, NetWalk \cite{yu2018netwalk}, DynamicNet \cite{zhu2012hybrid}, Method in \cite{yao2016link}  \\ \hline
GNN & Static & RecGNN \cite{scarselli2008graph}, GGNN \cite{li2016gated}, IGNN \cite{gu2020implicit}, Spectral Network \cite{bruna2013spectral}, GCN \cite{kipf2016semi}, GraphSAGE \cite{hamilton2017inductive}, DGN \cite{beaini2021directional}, ElasticGNN \cite{liu2021elastic}, SGC \cite{wu2019simplifying}, GAT \cite{velivckovic2017graph}, MAGNA \cite{wang2020multi}, MPNN \cite{gilmer2017neural}, GN block \cite{battaglia2018relational}, GNN-FiLM \cite{brockschmidt2020gnn}, GRNF \cite{zambon2020graph}, EGNN \cite{satorras2021n}, BGNN \cite{zhu2021bilinear}, MuchGNN \cite{zhou2021multi}, TinyGNN \cite{yan2020tinygnn}, GIN \cite{xu2018powerful}, RP-GNN \cite{murphy2019relational}, k-GNN \cite{morris2019weisfeiler}, PPGN \cite{maron2019provably}, Ring-GNN \cite{chen2019equivalence}, F-GNN \cite{azizian2020expressive}, DEGNN \cite{li2020distance}, GNNML \cite{balcilar2021breaking}, rGIN \cite{sato2021random}, DropGNN \cite{papp2021dropgnn}, PEG \cite{wang2021equivariant}, GraphSNN \cite{wijesinghe2021new}, NGNN \cite{zhang2021nested}, ID-GNN \cite{you2021identity}, CLIP \cite{dasoulas2021coloring}, APPNP \cite{klicpera2018predict}, JKNET \cite{xu2018representation}, GCN-PN \cite{zhao2019pairnorm}, DropEdge \cite{rong2019dropedge}, DGN-GNN \cite{zhou2020towards}, GRAND \cite{feng2020graph}, GCNII \cite{chen2020simple}, GDC \cite{hasanzadeh2020bayesian}, PDE-GCN \cite{eliasof2021pde}, SHADOW-SAGE \cite{zeng2021decoupling}, ClusterGCN \cite{chiang2019cluster}, FastGCN \cite{chen2018fastgcn}, LADIES \cite{zou2019layer}, GraphSAINT \cite{zeng2019graphsaint}, VR-GCN \cite{chen2018stochastic}, GBP \cite{chen2020scalable}, RevGNN \cite{li2021training}, VQ-GNN \cite{ding2021vq}, BNS \cite{yao2021blocking}, GLT \cite{chen2021unified}, H2GCN \cite{zhu2020beyond}, GPR-GNN \cite{chien2020adaptive}, WRGNN \cite{suresh2021breaking}, DMP \cite{yang2021diverse}, CPGNN \cite{zhu2021graph}, U-GCN \cite{jin2021universal}, NLGNN \cite{liu2021non}, GPNN \cite{yang2022graph}, HOG-GCN \cite{wang2022powerful}, Polar-GNN \cite{fang2022polarized}, GBK-GNN \cite{du2022gbk}, Geom-GCN \cite{pei2019geom}, GSN \cite{bouritsas2022improving}, MPSN \cite{bodnar2021weisfeiler}, GraphSTONE \cite{long2020graph}, DeepLPR \cite{chen2020can}, GSKN \cite{long2021theoretically}, SUBGNN \cite{alsentzer2020subgraph}, DIFFPOOL \cite{ying2018hierarchical}, PATCHY-SAN \cite{niepert2016learning}, SEAL \cite{zhang2018link}, DGCNN \cite{zhang2018end}, AGCN \cite{li2018adaptive}, DGCN \cite{zhuang2018dual}, CFANE \cite{pan2021unsupervised}, AdaGNN \cite{dong2021adagnn}, MCN \cite{lee2019graph}, Method in \cite{li2021dimensionwise}   \\ \cline{2-3} 
 & Spatial-temporal & GCRN \cite{seo2018structured}, Graph WaveNet \cite{wu2019graph}, SFTGNN \cite{li2021spatial}, CoST-Net \cite{ye2019co}, DSTN \cite{ouyang2019deep}, LightNet \cite{geng2019lightnet}, DSAN \cite{lin2020preserving}, H-STGCN \cite{dai2020hybrid}, DMSTGCN \cite{han2021dynamic}, PredRNN \cite{wang2017predrnn}, Conv-TT-LSTM \cite{su2020convolutional}, ST-ResNet \cite{zhang2017deep}, STDN \cite{yao2019revisiting}, ASTGCN \cite{guo2019attention}, DGCNN \cite{diao2019dynamic}, DeepETA \cite{wu2019deepeta}, SA-ConvLSTM \cite{lin2020self}, STSGCN \cite{song2020spatial}, FC-GAGA \cite{oreshkin2021fc}, ST-GDN \cite{zhang2021traffic}, HST-LSTM \cite{kong2018hst}, STGCN \cite{yu2018spatio}, PCR \cite{yang2018spatio}, GSTNet \cite{fang2019gstnet}, STAR \cite{xu2019spatio}, ST-GRU \cite{liu2019spatio}, Tssrgcn \cite{chen2020tssrgcn}, Test-GCN \cite{ali2021test}, ASTCN \cite{zhang2021adaptive}, STP-UDGAT \cite{lim2020stp}, STAG-GCN \cite{lu2020spatiotemporal}, ST-GRAT \cite{park2020st}, ST-CGA \cite{zhang2020spatial}, STC-GNN \cite{wang2021spatio}, STEF-Net \cite{liang2019deep}, FGST \cite{yi2021fine}, PDSTN \cite{miao2021predicting}, STAN \cite{luo2021stan}, GraphSleepNet \cite{jia2020graphsleepnet}, DCRNN \cite{li2018diffusion}, CausalGNN \cite{wang2022causalgnn}, SLCNN \cite{zhang2020spatio}, MRes-RGNN \cite{chen2019gated}, Method in \cite{wang2020traffic} \\ \cline{2-3} 
 & Dynamic & DyGNN \cite{ma2020streaming}, EvolveGCN \cite{pareja2020evolvegcn}, TGAT \cite{xu2020inductive}, CAW-N \cite{wang2020inductive}, DySAT \cite{sankar2018dynamic}, EHNA \cite{huang2020temporal}, TGN \cite{rossi2020temporal}, MTSN \cite{liu2021motif}, SDG \cite{fu2021sdg}, VGRNN \cite{hajiramezanali2019variational}, MNCI \cite{liu2021inductive}, FeatureNorm \cite{yang2020featurenorm} \\ \hline
\end{tabular}
\end{table}

\section{Traditional Graph Embedding}
\label{traditional}
The first category of graph embedding methods are traditional graph embedding methods. These methods map the nodes into the lower dimensions using different approaches such as random walks, factorization methods, and temporal point processes. We review these methods in static and dynamic settings in this section. Figure \ref{cat} shows the categories of static and dynamic traditional embedding methods. The upper part of Table \ref{table:methods} lists all the methods that we survey in this category.

\subsection{Traditional Static Graph Embedding}
\label{sec:second}
The traditional static graph embedding methods are developed for static graphs. The static graphs do not change over time and have a fixed set of nodes and edges. 
Graph embedding methods preserve different properties of nodes and edges in graphs such as node proximities. Here, we define first-order and second-order proximities. Higher order of proximities can be similarly defined.

\begin{definition}
(First-order proximity). Nodes that are connected with an edge have first-order proximity. Edge weights are the first-order proximity measures between nodes. Higher weights for edges show more similarity between two nodes connected by the edges. 

\end{definition}

\begin{definition}
(Second-order proximity). The second-order proximity between two nodes is the similarity between their neighborhood structures. Nodes sharing more neighbors are assumed to be more similar.

\end{definition}

 The traditional static graph embedding methods can be categorized into three categories: {\em factorization based}, {\em random walk based} and {\em non-GNN based deep learning} methods \cite{goyal2018graph,hamilton2017representation}. Below, we review these methods and the techniques they used.

\subsubsection{Factorization based}
Matrix factorization methods are the early works in graph representation learning. These methods can be summarized in two steps \cite{yang2017fast}. In the first step, a proximity-based matrix is constructed for the graph where each element of the matrix denoted as $P_{ij}$ is a proximity measure between two nodes $i,j$. Then, a dimension reduction technique is applied in the matrix to generate the node embeddings in the second step. 
In the Graph Factorization algorithm \cite{ahmed2013distributed}, the adjacency matrix is used as the proximity measure and the general form of the optimization function is as follows:
\begin{equation}
    \min_{z_i,z_j} \sum_{v_i, v_j \in V }|z_i^Tz_j -a_{ij}|
\end{equation}
where $z_i$ and $z_j$ are the node representation vectors for node $v_i$ and $v_j$. $a_{ij}$ is the element in the adjacency matrix corresponding to nodes $v_i$ and $v_j$. In GraRep \cite{cao2015grarep} and HOPE \cite{ou2016asymmetric}, the value of $a_{ij}$ is replaced with other measures of similarity including higher orders of adjacency matrix, Katz index \cite{katz1953new}, Rooted page rank \cite{song2009scalable} and the number of common neighbors. STRAP \cite{yin2019scalable} employs the personalized page rank as the proximity measure and approximates the pairwise proximity measures between nodes to lower the computation cost. In \cite{yang2017fast}, a network embedding update algorithm is introduced to approximately compute the higher order proximities between node pairs. In \cite{zhang2021learning}, it is suggested that using the same proximity matrix for learning node representations may limit the representation power of the matrix factorization based methods. Therefore, it generates node representations in a framework that learns the proximity measures and SVD decomposition parameters in an end-to-end fashion. Methods in  \cite{qiu2018network,zhang2018arbitrary,nie2017unsupervised,shen2018discrete,zhang2018billion,yang2018binarized,jiang2018spectral,huang2017accelerated,zhu2021refine,wang2017community,zhang2021learning,qiu2019netsmf,epasto2019single,brochier2019global,rossi2020structural,liu2019general,yang2017fast,lai2017prune,huang2020attributed,huang2017label,li2018community} are other examples of factorization based methods.

\subsubsection{Random walk based}
Random walk based methods have attracted a lot of attention because of their success in graph representation. The main concept that these methods utilize is generating random walks for each node in the graph to capture the structure of the graph and output similar node embedding vectors for nodes that occur in the same random walks. Using co-occurrence in a random walk as a measure of similarity of nodes is more flexible than fixed proximity measures in earlier works and showed promising performance in different applications. 


\begin{definition}
(Random walk). In a graph $G=(V,E)$, a random walk is a sequence of nodes $v_0,v_1,...,v_k$ that starts from node $v_0$. $(v_i,v_{i+1}) \in E$ and $k+1$ is the length of the walk. Next node in the sequence is selected based on a probabilistic distribution. 
\end{definition}


 DeepWalk \cite{perozzi2014deepwalk} and Node2vec \cite{grover2016node2vec} are based on the Word2vec embedding method \cite{mikolov2013efficient} in natural language processing (NLP). Word2vec is based on the observation that words that co-occur in the same sentence many times have a similar meaning. Node2vec and DeepWalk extend this assumption for graphs by considering that nodes that co-occur in random walks are similar. Therefore, these methods generate similar node embedding vectors for neighbor nodes. The algorithm of these two methods consists of two parts. In the first part, a set of random walks are generated, and in the second part, the random walks are used in the training of a SkipGram model to generate the embedding vectors. The difference between DeepWalk and Node2vec is in the way that they generate random walks. DeepWalk selects the next node in the random walk uniformly from the neighbor nodes of the previous node. Node2vec applies a more effective approach to generating random walks. In this section, we first explain the Node2vec random walk generation and then the SkipGram.

\begin{enumerate}
    \item \textit{Random Walk Generation.} Assume that we want to generate a random walk $v_0,v_1,...,v_k$ where $v_i \in V$. Given that the edge $(v_{i-1}, v_i)$ is already passed, the next node $v_{i+1}$ in the walk is selected based on the following probability:

\begin{equation}
     P(v_{i+1}|v_i) = \begin{cases}
      \frac{\alpha_{v_iv_{i+1}}}{Z} & \text{if}\ (v_{i+1},v_i) \in E \\
      0 & \text{otherwise}
    \end{cases}
\end{equation}
where Z is a normalization factor and $\alpha_{v_iv_{i+1}}$ is defined as:

\begin{equation}
     \alpha_{v_iv_{i+1}} = \begin{cases}
      1/p & \text{if}\ d_{v_{i-1}v_{i+1}} = 0 \\
      1 & \text{if}\ d_{v_{i-1}v_{i+1}} = 1 \\
      1/q & \text{if}\ d_{v_{i-1}v_{i+1}} = 2
    \end{cases}
\end{equation}
where $d_{v_{i-1}v_{i+1}}$ is the length of the shortest path between nodes $v_{i-1}$ and $v_{i+1}$ and takes values from $\{0, 1, 2\}$. The parameters $p$ and $q$ guide the direction of the random walk and can be set by the user. A large value for parameter $p$ encourages global exploration of the graph and avoids returning to the nodes that are already visited. A large value for $q$ on the other hand biases the walk toward local exploration. With the use of these parameters, Node2vec creates a random walk that is a combination of breadth-first search (BFS) and depth-first search (DFS). 
\item \textit{SkipGram.} After generating random walks, the walks are input to a SkipGram model to generate the node embeddings. SkipGram learns a language model, which maximizes the probability of sequences of words that exist in the training corpus. 
%
The objective function of SkipGram for node representation is: 
\begin{equation}
    \max_\Phi \sum_{v_i \in V} \log P(N(v_i)|\Phi(v_i))
\end{equation}
where $N(v_i)$ is the set of neighbors of node $v_i$ generated from the random walks. Assuming independency among the neighbor nodes, we have
\begin{equation}
     P(N(v_i)|\Phi(v_i)) = \prod_{v_k \in N(v_i)} P(\Phi(v_k)|\Phi(v_i))
\end{equation}
The conditional probability of $P(\Phi(v_k)|\Phi(v_i))$ is modeled using a softmax function:
\begin{equation}
     P(\Phi(v_k)|\Phi(v_i)) = \frac{\exp(\Phi(v_k)\Phi(v_i))}{\sum_{v_j \in V} \exp(\Phi(v_j)\Phi(v_i))}
\end{equation}
The softmax function nominator is the dot product of the node representation vectors. Since the dot product between two vectors measures their similarity, by maximizing the softmax function for neighbor nodes, the generated node representations for neighbor nodes tend to be similar. Computing the denominator of the conditional probability is time consuming between the target node and all the nodes in the graph. Therefore, DeepWalk and Node2vec approximate it using hierarchical softmax and negative sampling, respectively. 
\end{enumerate}
In HARP \cite{chen2018harp} a graph coarsening algorithm is introduced that generates a hierarchy of smaller graphs as $G_0, G_1,..., G_L$ such that $G_0=G$. Starting from the $G_L$ which is the smallest graph, the node embeddings that are generated for $G_i$ are used as initial values for nodes in $G_{i-1}$. This method avoids getting stuck in the local minimum for DeepWalk and Node2vec because it initializes the node embeddings with better values in the training process. The embedding at each step can be created using DeepWalk \cite{perozzi2014deepwalk} and Node2vec \cite{grover2016node2vec} methods. LINE \cite{tang2015line} is not based on random walks but because it is computationally related to DeepWalk and Node2vec, its results are usually compared with them. LINE generates node embeddings that preserve the first-order and second-order proximities in the graph using a loss function that consists of two parts. In the first part $L_1$, it minimizes the reverse of the dot product between connected nodes. In the second part $L_2$, for preserving the second-order proximity, it assumes that nodes that have many connections in common are similar. LINE trains two models that minimize $L_1$ and $L_2$ separately and then the embedding of a node is the concatenation of its embeddings from two models. Methods in \cite{wang2018united,xiao2020vertex,ribeiro2017struct,ivanov2018anonymous,chanpuriya2020infinitewalk,chanpuriya2020node,bhowmick2020louvainne,huang2021broader,zhou2017scalable,lyu2017enhancing,wang2020bringing} are some other variants of random walk based methods.

\subsubsection{Non-GNN based deep learning} 
SDNE \cite{wang2016structural} is based on an autoencoder which tries to reconstruct the adjacency matrix of a graph and captures nodes' first-order and second-order proximities. To that end, SDNE jointly optimizes a loss function that consists of two parts. The first part preserves the second-order proximity of the nodes and minimizes the following loss function:
\begin{equation}
    L_1 = \sum_{v_i \in V} |(x_i -x'_i) \odot b_i|
\end{equation}
where $x_i$ is the row corresponding to node $v_i$ in the graph adjacency matrix and $x'_i$ is the reconstruction of $x_i$. $b_i$ is a vector consisting of $b_{ij}$s for j from 1 to $n$ (the number of nodes in the graph). If $a_{ij} =0, b_{ij} = 1$; otherwise, $b_{ij} = \beta > 1$. $a_{ij}$ is the element corresponding to nodes $v_i$ and $v_j$ in the adjacency matrix. Using $b_i$, SDNE assigns more penalty for the error in the reconstruction of the non-zero elements in the adjacency matrix to avoid reconstructing only zero elements in sparse graphs. The second part capturs the first-order similarity and optimizes $L2$:
\begin{equation}
    L_2 = \sum_{(v_i,v_j) \in E} a_{ij}|(z_i -z_j)|
\end{equation}
where $z_i$ and $z_j$ are the embedding vectors for nodes $v_i$ and $v_j$, respectively. In this way, a higher penalty is assigned if the difference between the embedding vectors of two nodes connected by an edge is higher, resulting in similar embedding vectors for nodes connecting with an edge. This loss is based on ideas from Laplacian Eigenmaps \cite{belkin2001laplacian}. SDNE jointly optimizes $L_1$ and $L_2$ to generate the node embedding vectors. The embedding method DNGR \cite{cao2016deep} is also very similar to SDNE with the difference that DNGR uses pointwise mutual information of two nodes co-occurring in random walks instead of the adjacency matrix values. 
VGAE \cite{kipf2016variational} is a variant of variational autoencoders \cite{kingma2013auto} on graph data. 
The variational graph encoder encodes the observed graph data including the adjacency matrix and node attributes into low-dimensional latent variables.
\begin{align*}
q(Z|A,X) &= \prod_{i=1}^{N} q(z_i|A, X), \\
with \hspace{0.1cm} q(z_i|A,X) &= N(z_i|\mu_i, diag(\sigma_i^2))
\end{align*}
where $z_i$ is the embedding vector for node $v_i$, $\mu_i$ is a mean vector
and $\sigma_i$ is the log standard deviation vector of node $v_i$.
$A$ and $X$ are the adjacency matrix and attribute matrix of the graph, respectively. The variational graph decoder decodes the latent variables into the distribution of the observed graph data as follows:
\begin{align*}
 p(A|Z) &= \prod_{i=1}^{N}\prod_{j=1}^{N} p(a_{i,j}| z_i, z_j ), \\
with \hspace{0.1cm} p (a_{i,j} = 1| z_i, z_j ) &= sigmoid(z_i^T,z_j )
\end{align*}

The model generates embedding vectors that minimize the distance between the $p$ and $q$ probability distributions using the KL-divergence measure, SGD and reparametrization trick. Other works in \cite{tsitsulin2018verse,gao2018deep,al2019ddgk,meng2019co,chen2019fast,abu2018watch,du2018galaxy,he2020sneq}, also learn node embeddings using non-GNN based deep learning models.

\subsection{Traditional Dynamic Graph Embedding}
Most real-world graphs are dynamic and evolve, with nodes and edges added and deleted from them. Dynamic graphs are represented in two ways in the dynamic graph embedding studies: discrete-time and continuous-time.

\begin{definition}
(Discrete-time dynamic graphs). In discrete-time dynamic graph modeling, dynamic graphs are considered a sequence of graphs' snapshots at consecutive time points. Formally, dynamic graphs are represented as $G = G_0, G_1, ..., G_T$ which $G_i$ is a snapshot of the graph $G$ at timestamp $i$. The dynamic graph is divided into graph snapshots using a time granularity such as hours, days, months and years depending on the dataset and applications. 

\end{definition}


\begin{definition}
(Continuous-time dynamic graphs). In continuous-time dynamic graph modeling, the time is continuous, and the dynamic graph can be represented as a sequence of edges over time. The dynamic graph can also be modeled as a sequence of events, where events are the changes in the dynamic graphs, such as adding/deleting edges/nodes.

\end{definition}


\begin{definition}
(Dynamic graph embedding). We can use either the discrete-time or the continuous-time approach for representing a dynamic graph. Let $G_t=(V_t,E_t)$ be the graph at time $t$ with $V_t, E_t$ as the nodes and edges of the graph. Dynamic graph embedding methods map nodes in the graph to a lower dimensional space $d$ such that $d << |V|$.

\end{definition}

Dynamic graph embedding methods are more challenging than static graph embedding methods because of the challenges in modeling the evolution of graphs. Different methods have been proposed for dynamic graph embedding recently. Here, we provide an overview of the dynamic embedding methods and categorize these methods into four categories: Aggregation based, Random walk based, Non-GNN based deep learning and Temporal point process based \cite{kazemi2020representation,barros2021survey}.

\subsubsection{Aggregation based}
Aggregation based dynamic graph embedding methods aggregate the dynamic information of graphs to generate embeddings for dynamic graphs. These methods can fall into two groups:

\textit{1) Aggregating the temporal features.}
In these methods, the evolution of the graph is simply collapsed into a single graph and the static graph embedding methods are applied on the single graph to generate the embeddings. For example, the aggregation of the graph over time could be the sum of the adjacency matrices for discrete-time dynamic graphs \cite{liben2007link} or the weighted sum which gives more weights to recent graphs \cite{sharan2008temporal}. One drawback of these methods is that they lose the time information of graphs that reveals the dynamics of graphs over time. For instance, there is no information about when any edge was created. 
\textit{Factorization-based models} can also fit into the aggregation based category. The reason is that factorization based models save the sequence of graphs over time in a three dimensional tensor $\in \R ^{|V| \times |V| \times T}$ ($T$ is a time dimension) and then apply factorization on this tensor to generate the dynamic graph embeddings \cite{dunlavy2011temporal,liu2020dynamic,zhang2018timers,li2017attributed}.

\textit{2) Aggregating the static embeddings.}
These aggregation methods first apply static embedding methods on each graph snapshot in the dynamic graph sequence. Then, these embeddings are aggregated into a single embedding matrix for all the nodes in the graph. These methods usually aggregate the node embeddngs by considering a decay factor that assigns a lower weight to older graphs \cite{zhu2012hybrid,yao2016link}. In another type of these methods, the sequence of graphs from time $0$ to $t-1$ are fit into a time-series model like ARIMA that predicts the embedding of the next graph at time $t+1$ \cite{da2012time}. 

\subsubsection{Random walk based}
Random walk based approaches extend the concept of random walks in the static graphs for dynamic graphs. Random walks in dynamic graphs capture the time dependencies between graphs over time in addition to the topological structure of each of the graph snapshots. Depending on the definition of random walks, different methods include the temporal information of the graphs differently. 
CTDNE \cite{nguyen2018continuous} defines a temporal walk to capture time dependencies between nodes in dynamic graphs. CTDNE considers a continuous-time dynamic graph such as graph  $G=(V, E_T, T)$ which $V,E_T,T$ are nodes and edges of the graph and time $T: E \rightarrow \R^+$. Each edge $e$ in this graph is represented by a tuple $(u, v, t)$ which $u,v$ are the nodes connected by the edge and $t$ is the time of occurrence of that edge.

\begin{definition}
(Temporal walk). A temporal walk is a sequence of nodes $v_0, v_1,..., v_k$ such that $(v_i,v_{i+1}) \in E_T$ and $t_{(v_{i-1},v_i)} \le t_{(v_i,v_{i+1})}$.

\end{definition}

An important concept in CTDNE is that time is respected in selecting the next edge in a temporal walk. In order to generate these temporal walks, first a time and a particular edge in that time, $e = (u,v,t_e)$ is selected based on one of three probability distributions: uniform, exponential and linear. The uniform probability for an edge $e$ is $p(e) = 1/|E_T|$. The exponential probability is:
\begin{equation}
    p(e) = \frac{exp(t_e-t_{min})}{\sum_{e' \in E_T} exp(t_e'-t_{min})}
\end{equation}

where $t_{min}$ is the minimum time of an edge in the graph. Using exponential probability distribution, edges that appear at a later time are more likely to be selected. After selecting $e = (u,v,t_e)$, the next node in the temporal walk is selected from the neighbors of node $v$ in time $t_e+k$ where $k > 0$ again using one of the uniform, exponential or linear probability distributions. The generated temporal walks are then input to a SkipGram model and the temporal node representation vectors are generated.

DynNode2vec \cite{mahdavi2018dynnode2vec} is a dynamic version of Node2vec \cite{grover2016node2vec} and uses a discrete-time approach for dynamic graph representation learning. This method represents the dynamic graph as a sequence of graph snapshots over time as $G_0,G_1,..., G_T$. The embedding for the graph at time 0, $G_0$ is computed by applying Node2vec on $G_0$. Then, for next time points, the SkipGram model of $G_{t+1}$ is initialized using node representations from $G_t$ for nodes that are common between consecutive time points. New nodes will be initialized randomly. In addition, dynnode2vec does not generate random walks at each time step $i$ from scratch. Instead, it uses random walks from the previous time $i-1$ and only updates the ones that need to be updated. 
This method has two advantages. First, it saves time because it does not generate all the walks in each step. Second, since the SkipGram model at time $t$ is initialized with weights from time $t-1$, embedding vectors of consecutive times are in the same embedding space, embedding vectors of nodes change smoothly over time and the model converges faster.
LSTM-Node2vec \cite{khoshraftar2019dynamic} captures both the static structure and evolving patterns in graphs using an LSTM autoencoder and a Node2vec model. The dynamic graph is represented as a sequence of snapshots over time as $G_0,G_1,..., G_T$. For each graph $G_i$ at time $t_i$, first, a set of temporal walks is generated for each node in the graph. Each temporal random walk of a node $v$ is represented as ${w_0, w_1,...,w_L}$ of length $L$ which $w_j$ is a neighbor of the node $v$ at time $t_j$ in graph $G_j$ and $t_j < t_{j+1}$. These temporal walks demonstrate changes in the neighborhood structure of the node before time $t_i$.
EvoNRL \cite{heidari2020evolving} focuses on maintaining a set of valid random walks for the graph at each time point so that the generated node embeddings using these random walks stay accurate. To that end, EvoNRL updates the existing random walks from previous time points instead of generating random walks from scratch. Specifically, it considers four cases of edge addition, edge deletion, node addition and node deletion for evolving graphs and updates the affected random walks accordingly. For instance, in the edge addition case, EvoNRL finds random walks containing the nodes which are connected by the updated link and updates those walks. However, updating random walks is time consuming, especially for large graphs. Therefore, EvoNRL proposed an indexing mechanism for fast retrieval of random walks. Other examples of these category include \cite{ma2021temporal,du2018dynamic,yu2018netwalk,beladev2020tdgraphembed}.

\subsubsection{Non-GNN based deep learning}
This type of dynamic graph embedding methods use deep learning models such as RNNs and autoencoders. 
DynGEM \cite{goyal2017dyngem} is based on the static deep learning based graph embedding method SDNE \cite{wang2016structural}. Let the dynamic graph be a sequence of graph snapshots $G_0, G_1,..., G_t$. The embeddings for graph $G_0$ are computed using a SDNE model. The embedding of $G_i$ is obtained by running a SDNE model on $G_i$ that is initialized with the embeddings from $G_{i-1}$. This initialization leads to generating node embeddings at consecutive time points that are in the same embedding space and can reflect the changes in the graph at consecutive times accurately. As the size of the graph can change over time, DynGEM uses Net2WiderNet and Net2DeeperNet to account for bigger graphs \cite{chen2015fast}. 
 Dyn-VGAE \cite{mahdavi2019dynamic} is a dynamic version of VGAE \cite{kipf2016variational}. The input to dyn-VGAE is the dynamic graph as a sequence of graph snapshots, $G_0,G_1,..., G_T$. At each time point, the embedding of the graph snapshot $G_i$ is obtained using VGAE. However, the loss of the model at time $t$ has two parts. The first part is related to VGAE loss and the second loss is a KL divergence measure that minimizes the difference between two distributions as follows:
\begin{equation}
    L_s^t = KL[q_t(Z_t|X_t, A_t)||N(Z_{t-1}, \sigma^2)]
\end{equation}
where $q_t(Z_t|X_t, A_t)$ is the distribution of latent vectors at time $t$ and $N(Z_{t-1},\sigma^2)$ is a normal distribution with mean $Z_{t-1}$ and standard deviation $\sigma$. This loss places the current latent vectors $Z_t$ near latent vectors of previous time point $Z_{t-1}$. 
The loss function of all the models for the graph at time 0 to $T$ are jointly trained. Therefore, the generated representation vectors preserve both the structure of the graph at each time point and evolutionary patterns obtained from previous time points. 
Dyngraph2vec \cite{goyal2020dyngraph2vec} generates embeddings at time $t$ using an autoencoder. This method inputs adjacency matrices of previous times $A_0, A_1,..., A_{t-1}$ to the encoder and using the decoder reconstructs the input and generates the embeddings at time $t$. Dyngraph2vec proposes several variants using a fully connected model or a RNN/LSTM model for the encoder and the decoder: dyngraph2vecAE, dyngraph2vecAERNN and dyngraph2vecRNN. Dyngraph2vecAE uses an autoencoder, dyngraph2vecAERNN is based on an LSTM autoencoder and dyngraph2vecAERNN has a LSTM enocoder and a fully connected decoder. Other examples of non-GNN based deep learning methods include \cite{qiu2020temporal,singer2019node}.

\subsubsection{Temporal point process based}
This class of the dynamic graph embedding methods assumes that the interaction between nodes for creating the graph structure is a stochastic process and models it using temporal point processes. 
HTNE \cite{zuo2018embedding} generates embeddings for dynamic graphs by modeling the neighborhood formation of nodes as a hawkes process. In a hawkes process modeling, the occurrence of an event at time $t$ is influenced by events that occur before time $t$ and a conditional intensity function characterizes this concept. Let $G = (V,E,A)$ be the temporal network which $V, E, A$ are the nodes, edges and events. Each edge $(v_i,v_j)$ in this graph is associated with a set of events $a_{ij}=\{a_1 \rightarrow a_2 \rightarrow ...\} \subset A$ where each $a_i$ is an event at time $i$. 

\begin{definition}
(Neighborhood Formation Sequence). A neighborhood formation sequence for a node $v_i$ is a series of neighborhood arrival events $\{v_i: (u_0, t_0) \rightarrow (u_1, t_1) ...\rightarrow (u_k, t_k)\}$ where $u_i$ is a neighbor of $v_i$ that occurs at time $t_i$.

\end{definition}

HTNE models the neighborhood formation for a node $v$ using the neighborhood formation sequence $H_v$. The probability that an edge forms between node $v$ and a target neighbor $u$ at time t is represented using the following formula:
\begin{equation}
    p(u|v, H_v) = \frac{\lambda_{u|v}(t)}{\sum_{u'} \lambda_{u'|v}(t)}
\end{equation}

\noindent where $\lambda_{y|x}(t)$ is defined:
\begin{equation}
    \lambda_{u|v}(t) = exp(\mu_{u,v} + \sum_{h,u} \alpha_{h,u} \kappa(t-t_h))
\end{equation}

\noindent $\lambda_{u|v}(t)$ is the conditional intensity function of a hawkes process which is the arrival rate of target neighbor $u$ for node $v$ at time $t$ given the previous neighborhood formation sequence. $\mu_{u,v}$ is a base rate of edge formation between $u,v$ and it is equal to $|z_u -z_v|$. $h$ is a historical neighbor of the node $v$ in the neighborhood formation sequence in a time before $t$. $\alpha_{h,u}$ is the degree that the historical neighbor $h$ is important for $u$ and it equals $|z_h - z_u|$. $\kappa(t-t_h)$ is a decay factor to control the intensity of influence of a historical node on $u$. HTNE generates embedding vectors that maximize the $\sum_{v \in V} \sum_{u \in H_v} p(u|v, H_v)$ for all the nodes using SGD and negative sampling to deal with a large number of computations in the denominator of the probability function.
DyRep \cite{trivedi2019dyrep} captures the dynamic of graphs using two temporal point process models. DyRep argues that in the evolution of a graph two types of events occur: communication and association. Communication events are related to node interactions and association events are the topological evolution and these events occur at different rates. For instance, in a social network, a communication event such as liking a post from someone happens much more frequently than an association event like creating a new friendship. DyRep represents these two events as two temporal point process models. 
MTNE \cite{huang20motif} is based on two concepts of triad motif evolution and the hawkes process. This method considers the evolution of graphs as the evolution of motifs in the graphs and models that evolution using a hawkes process. MTNE argues that a model such as HTNE based on neighborhood formation processes considers network evolution at edge and node levels and can not reflect network evolution very well. Therefore, MTNE models dynamics in a graph as a subgraph (motif) evolution process. M2DNE \cite{lu2019temporal} is another example of temporal point process based dynamic embedding methods.

\subsubsection{Other methods}
DynamicTriad \cite{zhou2018dynamic} generates dynamic graph embeddings by modeling the triad closure process, which is a fundamental process in the evolution of graphs.

\begin{definition}
(Triad closure process). Let $(v_i, v_j, v_k)$ be an open triad in the graph at time $t$ which means that there are two edges $(v_i,v_j)$ and $(v_j,v_k)$ in the graph but no edge exists between $v_i$ and $v_k$. It is likely that an edge forms between $v_i$ and $v_k$ at time $t+1$ because of the influence of node $v_j$ and closes the open triad.

\end{definition}
DynamicTriad computes the probability that an open triad $(v_i, v_j, v_k)$ evolves into a closed triad under the influence of $v_j$ at time $t$ as $p_{tr}^t(i,j,k)$. An open triad can evolve in two ways: 1) It becomes closed because of the influence of any one of the neighbors. 2) Stays open because no neighbor could influence the creation of the open link. These two evolution traces are reflected in DynamicTriad loss function by maximizing $(p_{tr}^t(i,j,k))^{\alpha_{ijk}} \times (1-p_{tr}^t(i,j,k))^{(1-\alpha_{ijk})}$ for open triad samples that close under the influence of a neighbor and $1-p_{tr}^t(i,j,k)$ for those samples that do not close. $\alpha_{ijk}=1$ if an open triad closes at time $t+1$. The loss function also utilizes social homophily and temporal smoothness regularizations. \textit{Social homophily smoothness} assumes that nodes that are highly connected are more similar and should have similar embeddings. \textit{Temporal smoothness} assumes that network evolves smoothly and therefore, the distance between embeddings of a node at consecutive times should be small.

\section{GNN based graph embedding}
\label{GNN}
Graph Neural Net (GNN) based graph embedding methods are the second category of graph embedding methods, which employ GNNs to generate embeddings. These methods are different from traditional methods in that the GNN-based methods generalize well to unseen nodes. In addition, they can better take advantage of node/edge attributes. Table \ref{procon}  shows the advantages and disadvantages of the different categories of graph embedding methods.
 In this section, we first introduce GNNs. Then, three categories of GNN-based methods including static, spatial-temporal and dynamic GNNs (see Figure \ref{cat} for subcategories and Table \ref{table:methods} for the list of methods in each category) and their real-world applications are surveyed. Finally, we summarize the limitations of GNNs and the proposed solution to these limitations.




\subsection{Introduction to GNNs}
A GNN is a deep learning model which generates a node embedding by aggregating the node's neighbors embeddings. The GNN's intuition is that a node's state is influenced by its interactions with its neighbors in the graph. Below, we explain GNN's basic architecture and training.

\begin{table}[]
\caption{Comparison of traditional and GNN-based graph representation learning.}
\label{procon}
\begin{tabular}{lp{6cm}p{6.2cm}}
\hline
\textbf{Category} & \textbf{Advantages} & \textbf{Disadvantages} \\ \hline
Traditional & Higher expressive power, scalable in some categories &Not generalizable to unseen nodes, not    considering node/edge attributes easily \\
GNN-based& Generalize to unseen nodes, consider node/edge attributes, can do both task-specific and node similarity based training  & Expressive power, scalability, over-smoothing, over-squashing, homophily assumption and catastrophic forgetting. (More details in Section \ref{limitation}) \\ \hline
\end{tabular}
\end{table}

\subsubsection{Basic architecture}
GNNs can generate node representation vectors by stacking several GNN layers. Let $h_i^l$ represent the node embeddings for node $i$ at layer $l$. Each GNN layer takes as input the nodes embeddings. The node representations for node $i$ at each layer $l+1$ are updated using the following formula:


\begin{equation}
   h_i^{(l+1)}=f(h_i^l, \sum_{j \in N(i)} g(i,j)) 
\end{equation}
where $f$ and $g$ are learnable functions and $N(i)$ are the neighbors of node $i$. $h_i^0$ are the node $i$ initial features. At each layer, the embedding of the node $i$ is obtained by aggregating the embeddings of the node's neighbors. After passing through $L$ GNN layers, the final representation of node $i$ is $h_i^L$, which is the aggregation the node's neighbors of $L$ hops away from the node. 

\subsubsection{GNN training}
GNNs can be trained in supervised, semi-supervised and unsupervised frameworks. In supervised and semi-supervised frameworks, different prediction tasks focusing on nodes, edges and graphs can be employed for training the model. Here, we describe the other layers stacked after GNN layers to generate the prediction results.

\begin{itemize}
    \item Node-focused: For node-level prediction such as node classification, the GNN layers output node representations and then using an MLP or a softmax layer the prediction output is generated.
    
    \item Edge-focused: In edge-focused prediction including link prediction, given two nodes' representations, a similarity function or an MLP is used for the prediction task.
    
    \item Graph-focused:In graph-focused tasks such as graph classification, a graph representation is often generated by applying a readout layer on node representations. The readout function can be a pooling operation that aggregates representations of a graph's nodes to generate the graph representation vector. A clique pooling operation has also been proposed, which aggregates a graph's cliques for generating the graph embedding
 \cite{luzhnica2019clique}.

\end{itemize}


\begin{figure}
    \centering
    \includegraphics[scale=0.5]{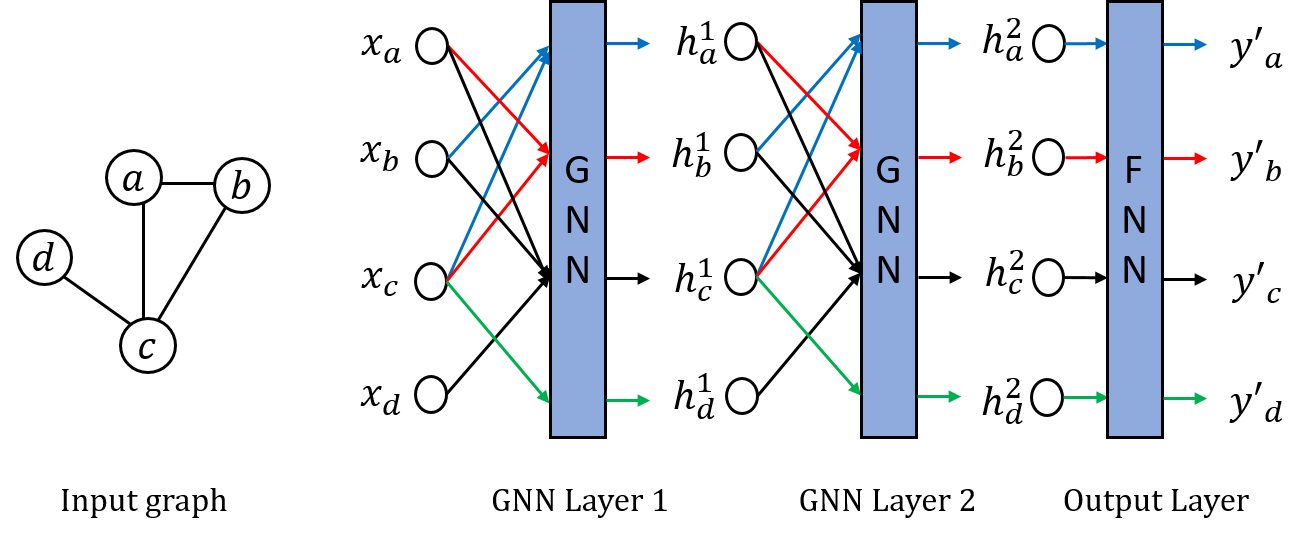}
    \caption{ A general supervised framework for training Graph neural net layers. Two GNN layers are applied on an input graph to compute the node representation vectors for its nodes. The colors on arrows show neighbors of a target node that are aggregated to generate the target node representation. $x_a$ is the feature vector of node $a$ and $h_a^1$ and $h_a^2$ are the representation vectors generated for the node $a$ after applying the first and second GNN layers.
    The generated embeddings are used in a node classification task. $y'_a$ is the predicted label for the node $a$.}
    \label{fig9}
\end{figure}


A typical way to train a GNN in a node classification task is by applying the cross entropy loss function as follows:
\begin{equation}
    L = \sum_{i \in V_{train}} (y_i \text{log}(\sigma(h_i^T\theta)+(1-y_i) \text{log}(1-\sigma(h_i^T\theta))
\end{equation}
where $h_i$ is the embedding of node $u$, which is the output of the last layer of GNN and $y_u$ is the true class label of the node and $\theta$ are the classification weights. Figure \ref{fig9} shows a general framework for training a GNN using a node classification task. 
There are three types of nodes in a node classification in GNN \cite{HamiltonBook}:
\begin{itemize}
    \item Training nodes: Nodes which their embeddings are computed in the last layer of GNNs and are included in the loss function computation.
    \item Transductive test nodes: Nodes which their embeddings are computed in the GNN but they are not included in the loss function computation.
    \item Inductive test nodes: They are not included in the GNN computation and loss function.
\end{itemize}
Transductive node classification in GNNs is equivalent to semi-supervised node classification. It refers to testing on transductive test nodes that are observed during training but their labels are not used. On the other hand, inductive node classification means that the testing is on inductive test nodes (unseen nodes) where these test nodes and all their adjacent edges are removed during training. The loss function for graph classification and link prediction tasks can be similarly defined using graph representations and pair wise node representations. In an unsupervised framework for GNN training, node similarities obtained from co-occurrence of nodes in the graph random walks can be used for model training. Graph neural nets often compute node representations using a graph-level implementation to avoid redundant computations for neighbors which are shared among nodes. In addition, formulating the message passing operations as matrix multiplications are computationally cheap. 
As an example for a basic GNN, the node embedding computation formula can be reformulated as:
\begin{equation}
    H^{(l+1)} = \sigma (\hat{A}H^lW^l)
\end{equation}
where $H^l$ contains the embedding of all the nodes in layer $l$ and $W_l$ is the weight matrix at layer $l$. $\hat{A}= \tilde{D}^{-1/2}\tilde{A}\tilde{D}^{-1/2}$, where $\tilde{A}=A+I_n$, $\tilde{D}_{ii}=\sum_j\tilde{A}_{ij}$, $I_n$ is an identity matrix and $A,D$ are the graph's adjacency and degree matrices. The graph-level implementation avoids redundant computations, however, it needs to operate on the whole graph which may lead to memory limitations. A number of methods have been proposed to alleviate the memory complexities of GNNs which will be discussed in Section \ref{limitation}.


\subsubsection{Other important concepts in GNNs}
In this section, we define some of the concepts that are frequently used in the GNN based graph representation literature.


\textbf{Receptive Field.}
The receptive field of a node in GNNs are the nodes that contribute to the final representation of the node. After passing through each layer of the GNN the receptive field of a node grows one step towards its distant neighbors.

\textbf{Graph Isomorphism.}
Two graphs are isomorphic if they have a similar topology. 
Some of the early works on GNN such as GCN \cite{kipf2016semi} and GraphSAGE\cite{hamilton2017inductive} fail to distinguish non-isomorphic graphs in some cases.

\textbf{Weisfeiler \& Lehman (WL) test.} The WL test \cite{leman1968reduction} is a classic algorithm for testing graph isomorphism. It has been shown that the representation power of the message passing GNNs is upper bounded by this test \cite{xu2018powerful}. The WL test successfully determines isomorphism between different graphs but there are some corner cases that it fails. Similarly, GNNs fails in those cases. The simple way of thinking about how this test works is that it first counts the number of nodes in two graphs. If two graphs have a different number of nodes, they are different. If two graphs have a similar number of nodes, it checks the number of immediate neighbors of each node. If the number of immediate neighbors of each node is the same, it goes to check the second-hop neighbors of nodes. If two graphs were similar in all these cases, then they are identical or isomorphic. 

\textbf{Skip connections.}
A skip connection in deep architectures means skipping some layers in the neural network and feeding one layer's output as an input to the next layers, not just the immediate next layer. An skip connection helps in alleviating the vanishing gradient effect and preserving information from previous layers. For instance, skip connections are used in GraphSAGE \cite{hamilton2017inductive} update step. This method concatenates the node representation at the previous level with the aggregated representation from node neighbors from the previous layer in the update step. This way, it preserves more node-level information in the message passing.

\subsection{Static Graph Neural Nets} Static GNN based graph embedding methods are suitable for graph representation learning on static graphs, which do not change over time.  These methods can be divided into two classes: Recurrent GNNs and Convolutional GNNs that will be explained below.

\subsubsection{Recurrent Graph Neural Net (RecGNN)}
RecGNNs are the early works on GNN that are based on RNNs. The original GNN model proposed by Scarselli et al. \cite{scarselli2008graph} used the assumption that nodes in a graph constantly exchange information until they reach an equilibrium. In this method, the representation of node $v$ at iteration $t$, $h_v^t$ is defined using the following recurrence equation:

\begin{align}
    h_v^t &=  \sum_{u \in N(v)} f(x_v,x_{(v,u)}^e, h_u^{(t-1)}, x_u)
\end{align}

\noindent where $f$ is a recurrent function. $N(v)$ is a set of neighborhood nodes of node $v$. $x_v, x_u$ are feature vectors of nodes $v,u$ and $x_{(v,u)}^e$ is the feature vector of the edge $(u,v)$.  This GNN model recursively runs until convergence to a fixed point. Therefore, the final representation $h_v^T$ in this method is a vector that $h_v^T=f(h_v^{T-1})$. 
In this model, $h_v^0$ is initialized randomly. The initialization of node representation vectors does not matter in this model because the function $f$ recursively converges to the fixed point using any value as an initialization. For learning the model parameters, the states $h_v^t$ are iteratively computed until the iteration $T$. An approximate fixed point solution is obtained and used in a loss function to compute the gradients. This model has several limitations. One limitation is that if $T$ is large, the iterative computation of node representation until convergence is time consuming. Furthermore, the node representations obtained from this model are more suitable for graph representation than node representation as the outputs are very smooth.
GGNN \cite{li2015gated} uses gated recurrent unit (GRU)  as the recurrent function in the original RecGNN method proposed by Scarselli et al. \cite{scarselli2008graph}. The advantage of using GRU is that the number of recurrence steps is fixed and the aggregation does not need to continue until convergence. The $h_v^t$ formula is as follows:
\begin{equation}
    h_v^t = GRU(h_v^{(t-1)}, \sum_{u \in N(v)} h_u^{(t-1)})
\end{equation}
The $h_v^0$ are initialized with node features. Implicit Graph Neural Net (IGNN) \cite{gu2020implicit} is another recurrent GNN that generates node representations by iterating until convergence with no limit on the number of neighbor hops. However, it guarantees the existence of the solution for the equilibrium equations by defining the concept of well-posedness for GNNs, which was previously defined for neural nets \cite{el2021implicit} and enforcing it at the training time.

\subsubsection{Convolutional Graph Neural Net (ConvGNN)}
ConvGNNs are a well-known category of graph neural nets. These methods generate node embeddings using the concept of convolution in graphs. The difference between ConvGNNs and RecGNNs is that ConvGNNs use CNN based layers to extract node embeddings instead of RNN layers in RecGNNs. 
There are three key characteristics in CNNs that make them attractive in graph representations. 1) Local connections: CNN can extract local information from neighbors for each node in the graph,  2) shared weights: weight sharing in node representation generates node embeddings that consider the information of other nodes in the graph, 3) multiple layers: each layer of convolution can explore a layer of proximities between nodes \cite{zhou2020graph}. 
ConvGNNs have two categories that can overlap: Spectral based and Spatial based methods. The spectral based methods have roots in graph signal processing and define graph signal filters. The spatial based methods are based on information propagation and message passing concepts from RecGNNs and are more preferred than spectral methods because of efficiency and flexibility. Here, we explain these two categories in more detail.

\bigskip
\textit{1) Spectral based.}
Spectral based graph neural nets utilize mathematical concepts from graph signal processing. \textbf{Spectral Network} \cite{bruna2013spectral} is one of the early works that defines convolution operation on graphs. Here, we define some of the main concepts shared among spectral based GNNs.  

\begin{definition}
(Graph Signal). In graph signal processing, a graph signal $x \in \R^n$ is an array of $n$ real or complex values for $n$ nodes in the graph. 

\end{definition}

\begin{definition}
(Eigenvectors and eigenvalues (spectrum)). Let $L= D - A$ be the graph Laplacian of graph $G$ which $D, A$ are the graph's degree matrix and adjacency matrix. The normalized graph Laplacian matrix is $L_N = I_n-D^{-1/2}AD^{-1/2}$ which can be factorized as $L_N=U \Lambda U^T$. $U$ is the matrix of eigenvectors and $\Lambda$ is the diagonal matrix of ordered eigenvalues. The set of eigenvalues of a matrix are also called the spectrum of the matrix.

\end{definition}

\begin{definition}
(Graph Fourier transform). The graph fourier transform is $F=U^Tx$ which maps the graph signal $x$ to a space formed by the eigenvectors of $L_N$.  

\end{definition}

\begin{definition}
(Spectral Graph Convolution). The spectral graph convolution of the graph signal $x$ with a filter $g \in \R^n$ is defined  as:
\begin{gather}
    x * g = F^{-1}(F(x) \odot F(g)) = U(U^Tx \odot U^Tg) \\
   \Rightarrow x *g_\theta = Ug_\theta U^Tx  \\
   \text{ where } g_\theta = diag(U^T g)
\end{gather}
\end{definition}

Different spectral based ConvGNNs use a different graph convolution filter $g_\theta$. For instance, Spectral CNN \cite{bruna2013spectral} defines $g_\theta$ as a set of learnable parameters. One of the main limitations of this method is the eigenvalue decomposition computational complexity. This limitation was resolved by applying several approximations and simplification in future works.
Graph Convolutional Network (GCN) \cite{kipf2016semi} uses a layerwise propagation rule based on multiplying the first-order approximation of localized spectral convolution filter $g_\theta$ with a graph signal $x$ as follows:

\begin{equation}
    x * g_\theta = \theta(I_n+D^{-1/2}AD^{-1/2})x
\end{equation}
where $D, A$ are the degree matrix and adjacency matrix of a graph G and $I_n$ is an identity matrix with 1 on the diagonal and 0 elsewhere. $\theta$ represents the filter parameters. GCN also modifies the convolution operation into a layer defined as $H= X * g_\Theta= f(\bar{A}X\Theta)$, where $f$ is an activation function and $\bar{A} = I_n+D^{-1/2}AD^{-1/2}$. Using a renormalization trick $\bar{A}$ is replaced with $\hat{A}= \tilde{D}^{-1/2}\tilde{A}\tilde{D}^{-1/2}$, where $\tilde{A}=A+I_n$ and $\tilde{D}_{ii}=\sum_j\tilde{A}_{ij}$. Therefore, the formulation of $h_v^{l+1}$ for node $v$ becomes:
\begin{equation}
    h_v^{l+1} = f(\Theta^l (\sum_{u \in \{N(v) \cup v\}} \hat{A}_{v,u}x_u)) 
\end{equation}
where $\bar{A}$ is a constant and is approximately computed in a preprocessing step. $N(v)$ is the set of neighbors of a node $v$. Therefore, $h_v^{l+1}$ value can be roughly approximated as:

\begin{equation}
    h_v^{l+1} \approx f(\Theta^l . Mean( h_v^{l} \cup \{h_u^{l}, \forall u \in N(v)\} )) 
\end{equation}
In a neural net setting, $f$ is an activation function such as ReLU and $\Theta^l$ is the matrix of parameters at layer $l$. GCN can be viewed as a spatial-based GNN because it updates the node embeddings by aggregating information from neighbors of nodes. 
In \cite{li2021dimensionwise}, a spectral based model is proposed which jointly learns relations between nodes and relations between attributes of nodes. The node embeddings in this model are the output of a 2D spectral graph convolution defined as $Z=GXF$. In this formula, $X$ is a node feature matrix and $G$ and $F$ are an object graph convolutional filter and an attribute graph convolutional filter. The object graph convolutional filter is defined by designing a filter on the adjacency matrix of the graph. For defining the attribute graph convolutional filter, an attribute affinity graph is constructed on the original graph by applying either positive point-wise mutual information or word embedding based KNN on the attributes of the nodes. 
Directional Graph Networks (DGN) \cite{beaini2021directional} defines directions for information propagation in the graph using vector fields to improve the message passing in a specific direction in the current GNNs. In this method, the contribution of a neighbor node depends on its alignment with the receiving node's vector field. The vector fields denoted by $B$ are defined using the $k$ lowest frequency eigenvectors of the Laplacian matrix of the graph as they preserve the global structure of graphs \cite{grebenkov2013geometrical}. The node representations are obtained by multiplication of the matrix $B$ and the adjacency matrix of the input graph. In \cite{ma2021unified}, it has been shown that most common GNNs perform $l_2$-based graph smoothing on the graph signal in the message passing, which leads to global smoothness. Motivated by the trend filtering idea \cite{wang2015trend}, Elastic GNN \cite{liu2021elastic} accounts for different smoothness levels for different regions of the graph using $l_1$-based graph smoothing. In \cite{zheng2021framelets}, a framelet graph convolution is proposed. This method is based on graph framelets and their transforms \cite{zheng2022decimated}. Framelet convolution can lower the feature and structure noises in graph representation. This method decomposes the graph into low-pass and high-pass matrices and generates framelet coefficients. Then, the coefficients are compressed by shrinkage activation, which improves the network denoising properties. Simple Graph Convolution (SGC) \cite{wu2019simplifying} is a graph convolution network which simplifies the GCN model by removing the non-linear activation functions at consecutive layers. This study theoretically proves that this model corresponds to a fixed low-pass filtering in spectral domain in which similar nodes have similar embeddings. 
Many other studies introduce different variants of spectral-based GNNs \cite{miao2021degnn,ma2020path,klicpera2019diffusion,zhao2021adaptive,zhang2018end,li2018adaptive,zhuang2018dual,wang2022powerful}.

\bigskip
\textit{2) Spatial based.}
Spatial-based ConvGNNs define the graph convolution similar to applying CNN on images. Images can be viewed as a graph such that the nodes are the pixels and the edges are the proximity of pixels. When a convolution filter applies to an image, the weighted average of the pixel values of the central node and its neighbor nodes are computed. Similarly, the spatial-based graph convolutional filters generate a node representation by aggregating the node representations of neighbors of a node. One of the advantages of spatial-based ConvGNNs is that the learned parameters of models are based on close neighbors of nodes and therefore, can be applied on different graphs with some constraints. In contrast, spectral-based models learn filters that depend on eigenvalues of a graph Laplacian and are not directly applicable on graphs with different structures.
GraphSAGE \cite{hamilton2017inductive} is one of the early spatial based ConGNNs. This method generates node embeddings iteratively. The node embeddings are first initialized with node attributes. Then, a node embedding at iteration $k$ is computed by concatenating the aggregation of the node's neighbor and the node embedding at iteration $k-1$. For example, for a node $v$,
\begin{gather*}
    h_{N(v)}^k = Aggregate_k ({h_{u}^{k-1}, \forall u \in N(v)})
    \\
    h_{v}^k = \sigma (W^k.Concat(h_{v}^{k-1}, h_{N(v)}^k))
\end{gather*}

\noindent where $h^k$ and $W^k$ are the node embedding and weight matrix at iteration $k$. $N(v)$ is the set of neighbors of $v$. GraphSAGE leverages mean, LSTM and pooling aggregators as follows:
\begin{itemize}
\item Mean aggregator. The mean aggregation is similar to GCN \cite{kipf2016semi} which takes mean over neighbors of a node. The difference is that GCN includes the node representation $h_v^{k-1}$ in the mean but GraphSAGE concatenates the node representation with the mean aggregation of neighbor nodes. This way, GraphSAGE avoids node information loss.
\item LSTM aggregator. An LSTM aggregator aggregates neighbor nodes representations using an LSTM structure. It is important to note that LSTM preserves the order between nodes, however, there is no order among neighbor nodes. Therefore, GraphSAGE inputs a random permutation of nodes to alleviate this problem. 
\item Pooling aggregator. In this aggregation, each neighbor node is fed through a fully-connected neural net and then an elementwise max operation is applied on the transformed nodes as follows:
\begin{equation}
    Aggregate^{pool} = \text{max } (\{ \sigma (W_{pool} h_{u}+b), \forall u \in N(v)\})
\end{equation}
The above equation uses the max operator for pooling however mean operator can be used as well. The pooling aggregator is symmetric and learnable. The pooling aggregation intuition is that it captures different aspects of the neighborhood set of a node.
\end{itemize}
The aggregation continues until $K$ iterations. 
The model is trained using a loss function that generates similar node embeddings for nearby nodes in an unsupervised setting. The unsupervised loss can be replaced with task specific objective functions. 
Graph Attention Network (GAT) \cite{velivckovic2017graph} utilizes the self-attention mechanism \cite{vaswani2017attention} to generate node representations. Unlike GCN that assigns a fixed weight to neighbor nodes, GAT learns a weight for a neighbor depending on the importance of the neighbor node. The state of node $v$ at layer $k$ is formulated as follows:
\begin{align}
    h_v^k&=\sigma(\sum_{u \in N(v)}  
    \alpha_{vu}^k W^kh_u^{k-1}) \\
    \alpha_{vu} &=\frac{exp(\text{LeakyReLU}(a^T[Wh_u||Wh_v]))}{\sum_{k \in N_v}exp(\text{LeakyReLU}(a^T[Wh_u||Wh_k]))} \\
    h_v^{0} &=x_v
\end{align}
where $\alpha_{vu}$ is the attention coefficient of node $v$ to its neighbor $u$ defined using a softmax function. $N_v$ is the neighbor set of node $v$. $W$ is the weight matrix and $a$ is a weight vector. $||$ is the concatenation symbol.  In addition to self-attention, GAT's results benefit from using multi-head attention. Similar to GraphSAGE, GAT is trained in an end-to-end fashion and outputs node representations. 
Multi-hop Attention Graph Neural Network (MAGNA) \cite{wang2020multi} generalizes the attention mechanism in GAT \cite{velivckovic2017graph} by increasing the receptive fields of nodes in every layer. Stacking multiple layers of GAT has the same effect, however that causes the oversmoothing problem. MAGNA first computes the 1-hop attention matrix for every node and then uses the sum of powers of the attention matrix to account for multi-hop neighbors in every layer. To lower the computation cost, an approximated value for the multi-hop neighbor attention is computed. MAGNA model aggregates the node features with attention values and passes the values through a feed forward neural network to generate the node embeddings. 
Message Passing Neural Net (MPNN) \cite{gilmer2017neural} proposes a general framework for ConvGNNs. In MPNN framework, each node sends messages based on its states and updates its states based on messages received from its immediate neighbors. The forward pass of MPNN has two parts: A message passing and a readout phase. In the message passing phase, a message function is utilized for information propagation and the node state is updated as follows:
\begin{gather}
    h_v^t=U_t(h_v^{t-1}, \sum_{u \in N(v)} M_t(h_v^{t-1}, h_u^{t-1}, e_{vu})) \\
    h_v^{0} =x_v
\end{gather}
where $M_t$ is the message function and $U_t$ updates the node representation. $U_t, M_t$ are learnable functions. $e_{vu}$ is the information of an edge $(v,u)$. In the readout phase,  the readout layer generates the graph embeddings using the updated node representations, $h_G=R(h_v^t|v \in G)$. Different ConvGNN methods can be formulated using this framework using different functions for $U_t, M_t, R$. 
GN block \cite{battaglia2018relational} proposes another general framework for Graph neural nets that some of the GNN methods could fit in its description. A GN block learns nodes, edges and the graph representations denoted as $h_i^l, e_{ij}^l, u^l$ respectively. Each GN block contains three update functions, $\phi$ and three aggregation functions, $\rho$:
\begin{align}
    e_{ij}^{l+1} &= \phi^e(e_{ij}^l, h_i^l, h_j^l, u^l) && m_i^{l+1} =\rho^{e \rightarrow v} (\{e_{ij}^{l+1}, \forall j \in N(i)\})\\
    h_i^{l+1} &= \phi^v(m_i^{l+1}, h_i^l, u^l) && m_V^{l+1} =\rho^{v \rightarrow u} (\{h_i^{l+1}, \forall i \in V\})\\
    u^{(l+1)} &= \phi^u(m_E^{l+1}, m_V^{l+1}, h_i^l, u^l) && m_E^{l+1} =\rho^{e \rightarrow u} (\{e_{ij}^{l+1}, \forall (i,j) \in E\})
\end{align}

The GN assumption is that computation on a graph starts from an edge to a node and then to the entire graph. This phenomenon is formulated with update and aggregation functions: 1) $\phi^e$ updates the edge representations for each edge. 2) $\rho^{e \rightarrow v}$ aggregates the updated edge representations for the edges connected to each center node. 3) $\phi^v$ updates the node representations. 4) $\rho^{v \rightarrow u}$ aggregates node representation updates for all nodes. 5) $\rho^{e \rightarrow u}$ aggregates edge representation updates for all edges. 6) finally, the entire graph representation is updated by $\phi^u$. 
GNN-FiLM \cite{brockschmidt2020gnn} generates node embedding using the feature-wise linear modulation (FiLM) idea that was introduced in the visual question answering area \cite{perez2018film}. Many common GNNs such as GCN \cite{kipf2016semi} and GraphSAGE \cite{hamilton2017inductive} propagate information along edges using information from the source node of the edges. In GNN-FiLM, the target node representation transformation is computed and applied to incoming messages to generate the feature-wise modulation of the incoming messages. Graph Random Neural Features (GRNF) \cite{zambon2020graph} generates graph embeddings by preserving the metric structure of the graphs in the embedding space and therefore distinguishing between any pair of non-isomorphic graphs. This method is based on a family of graph neural feature maps. The graph neural feature maps are graph neural networks that can separate graphs. The outputs of these GNNs, which are scalar features, are then concatenated to generate the graph embedding. E(n) Equivariant Graph Neural Network (EGNN) \cite{satorras2021n} is a rotation, translation and permutation equivariant GNN. These properties are fundamental in representing structures that show rotation and translation symmetric characteristics, such as molecular structures \cite{ramakrishnan2014quantum}. EGNN takes as inputs a feature vector and an n-dimensional coordinates vector for each graph node along with the edge information and outputs the node embeddings. The main difference between this method and common GNNs is that the relative squared distance between a node's coordinates and neighbors has been considered in the GNN message passing operation. Bilinear Graph Neural Network (BGNN) \cite{zhu2021bilinear} argues that the neighbors of a node can have interactions that may affect the node representations. Therefore, it augments the aggregation of neighbors of a node by pairwise interactions of neighbor nodes. Motivated by factorization machines \cite{he2017neural} , it models the neighbors' interaction using a bilinear aggregator denoted by $BA$, which computes the average of pairwise multiplication of neighbor nodes of a node. Then, the convolution operator is defined as follows:
\begin{equation}
    H^{(k)} = (1-\alpha).AGG(H^{(k-1)}, A)+\alpha.BA(H^{(k-1)},A)
\end{equation}
where  $H^{(k)}$ is the node representation at $k$-th layer and $\alpha$ is a tradeoff parameter between two components. In \cite{zhang2020improving}, it is theoretically shown that all attention-based GNNs fail in distinguishing between certain structures due to ignoring the cardinality information in aggregation. Therefore, this paper introduces two cardinality-preserved attention (CPA) models named Additive and Scaled. The formulation of the Additive model is as follows:
\begin{equation}
    h_i^l = f^l(\sum_{j \in N(i)} \alpha_{ij}^{l-1}h_i^{l-1}+w^l\odot \sum_{j \in N(i)} h_i^{l-1})
\end{equation}
which the first term is the original attention formula and the second term captures the cardinality information. The Scaled model formula is:
\begin{equation}
    h_i^l = f^l(\psi^l(|N(i)|)\odot\sum_{j \in N(i)} \alpha_{ij}^{l-1}h_i^{l-1})
\end{equation}
where $\psi(|N(i)|)$ is a function that maps the cardinality value to a non-zero vector. Both these models improve the distinguishing power of the original attention model. Multi-Channel graph neural network (MuchGNN) \cite{zhou2021multi} generates graph representations using graph pooling operation. However, instead of shrinking the graph layer by layer using graph pooling which may result in loss of information, it shrinks the graph hierarchically. This method generates a series of graph channels at each layer and applies graph pooling on them to generate the graph representation at each layer. The final graph representation is the concatenation of graph representations at each layer. Policy-GNN \cite{lai2020policy} captures information for each node using different iterations of aggregations to capture the graph's structural information better. To that end, it uses meta-policy \cite{zha2019experience} trained by deep reinforcement learning to choose the number of aggregations per node. TinyGNN \cite{yan2020tinygnn} proposes a small GNN with a short inference time. In order to capture the local structure of the graph, this method generates node representations by aggregating peer-aware representations of the node's neighbors. Peer-aware representations consider the interactions between peer nodes, which are neighbor nodes with the same distance from the center node. In addition, inspired by knowledge distillation \cite{hinton2015distilling}, it proposes a neighbor distillation strategy (NDS) in a teacher-student network. The teacher network is a regular GNN and has access to the entire neighborhood and the student network is a small GNN that imitates the teacher network. Other spatial based convolution GNNs include \cite{liu2021tail,vignac2020building,zhang2020factor,nikolentzos2020random,xu2021automorphic,he2021learning,yun2021neo,pei2019geom,you2019position,zhu2021neural,lee2019graph,pan2021unsupervised,niepert2016learning,ying2018hierarchical}.

\subsection{Spatial-Temporal Graph Neural Net (STGNN)}
Spatial-temporal GNNs are a category of GNN that capture both the spatial and temporal properties of a graph. They model the dynamics of graphs considering the dependency between connected nodes. There are wide applications for STGNNs such as traffic flow forecasting \cite{wang2020traffic,li2021spatial,zhang2020spatio,chen2019gated,guo2019attention,zhang2021traffic}, epidemic forecasting \cite{wang2022causalgnn} and sleep stage classification \cite{jia2020graphsleepnet}. For instance, in traffic prediction, the future traffic in a road is predicted considering the traffic congestion of its connected roads in previous times. Most of the STGNN methods fall into CNN based and RNN based categories which integrate the graph convolution in CNNs and RNNs, respectively.

\subsubsection{RNN based}
Graph Convolutional Recurrent Network (GCRN) \cite{seo2018structured} is an example of RNN based STGNN. In this method, an LSTM network is combined with the convolution operation. GCRN has two variants. In the first variant, a CNN layer is stacked with an LSTM layer. The CNN layer extracts the features at time $t$ and the LSTM captures the temporal behavior of nodes over time.
In the second variant, GCRN replaces the matrix multiplication operation in the LSTM with the graph convolution operation. 
In \cite{wang2020traffic}, the spatial and temporal correlations between nodes are modeled using three components including a spatial graph neural network layer, a GRU layer and a transformer layer. The input to the model is a sequence of graphs over time. The spatial graph neural network layer captures the spatial relations between nodes in each graph. Then, a GRU and transformer layers are applied on the sequence of graphs which are output from previous layer and capture the temporal relations between graphs over time. Other RNN-based spatial-temporal GNNs include methods in  \cite{ding2022spatio,chen2019gated,li2018diffusion,wang2022causalgnn,ye2019co,geng2019lightnet,wang2017predrnn,su2020convolutional,yao2019revisiting,wu2019deepeta,lin2020self,kong2018hst,yang2018spatio,xu2019spatio,liu2019spatio,zhang2021adaptive,wang2021spatio,liang2019deep,yi2021fine,miao2021predicting,li2018diffusion}.


\subsubsection{CNN based} Graph WaveNet \cite{wu2019graph} is a CNN based spatio-temporal GNN. This method takes as input a graph and feature matrices of nodes for $m$ previous time steps and the goal is to predict the next $n$ feature matrices.  For example, in a traffic prediction application, a feature matrix is a $N \times d$ matrix that each row contains traffic features of a node. Each node is a sensor or a road. $N$ is the number of nodes and $d$ is the feature vector dimension. The framework of Graph WaveNet consists of two building blocks: a graph convolution layer and a temporal convolution layer. In the graph convolution layer, it combines a diffusion convolution layer \cite{li2018diffusion} with a self-adaptive adjacency matrix. The central part of the self-adaptive adjacency matrix is multiplying source and target node embeddings that are initialized randomly and are learned during the model training. The temporal convolution layer adopts a gated version of a dilated causal convolution network \cite{yu2015multi}. 
In Spatial-Temporal Fusion Graph Neural Networks (SFTGNN)\cite{li2021spatial}, 
instead of modeling the spatial and temporal correlations of nodes separately, a spatial-temporal fusion graph is constructed using three $N \times N$ matrices to capture three kinds of correlation for each node. 1) A spatial graph for spatial neighbors, 2) A temporal graph for nodes with similar temporal sequences and 3) A temporal connectivity graph for connection of a node at nearby time points. The spatial-temporal fusion graph is then input to spatial-temporal fusion graph neural module (STFGN module) that generates node representations. In \cite{zhang2020spatio,guo2019attention,zhang2021traffic,jia2020graphsleepnet,ouyang2019deep,lin2020preserving,dai2020hybrid,han2021dynamic,zhang2017deep,guo2019attention,diao2019dynamic,song2020spatial,oreshkin2021fc,zhang2021traffic,yu2018spatio,fang2019gstnet,chen2020tssrgcn,ali2021test,lim2020stp,lu2020spatiotemporal,park2020st,zhang2020spatial,luo2021stan}, many other CNN-based spatial-temporal GNNs are proposed.


\subsection{Dynamic Graph Neural Net (DGNN)}
Dynamic Graph Neural Nets (DGNN) are Graph neural nets that model a broad range of dynamic behaviors of a graph, including adding or deleting nodes and edges over time. 
EvolveGCN \cite{pareja2020evolvegcn} is a dynamic GNN method. 
The idea behind it is to use an RNN model to update weights of the GCN at each time point and capture dynamics of the graph. At each time step $t$, a GCN layer is used to represent the graph at time $t$. In order to integrate historical information of the nodes, the initial weights for the GCN at time $t$ are the hidden states/output of RNN based models which take as input the weights of previous GCN models. Each RNN based model is assigned to a separate layer of GCN models. 
The EvolveGCN model is trained end to end for link prediction and edge/node classification tasks.
DyGNN \cite{ma2020streaming} proposes a dynamic GNN method that consists of two components: Update and propagation components. These two components work in parallel to update and propagate the information of a new interaction in the graph. Let $(v_s, v_g, t)$ represent a new directed edge that emerges between a source node $v_s$ and a target node $v_t$ at time $t$. The update component updates the two nodes $v_s, v_t$ representations and the propagate component propagates the interaction information to \textit{influenced nodes} which are defined as 1-hop neighbors of the two interacting nodes. 
Dynamic Self-Attention Network (DySAT) \cite{sankar2018dynamic} consists of two components, a structural block followed by a temporal block to capture the structural and temporal properties of a graph. This method defines dynamic graphs as a series of graphs over time. In order to capture the structure of the graph at each time point, the structural block, which is a variant of GAT is applied to the graph at each time point.
Then, to capture dynamic patterns of nodes, DySAT applies a temporal self attention layer in the temporal block. The inputs to the temporal block are node representations overtime for every node $v$ such that the node representation $x_v^t$ attends over the historical representation of the node $(<t)$ to generate the final embedding of each node. 
Embedding via Historical Neighborhoods Aggregation (EHNA) \cite{huang2020temporal} 
aggregates the historical neighbors of a node to capture the evolution of the node in the graph. 
In order to capture the historical neighbors of a node, EHNA first generates $k$ temporal random walks for each node. The transition probability of each edge in a temporal random walk depends on the weight and time of the edge.
The generated random walks for each node $x$ are first aggregated using an LSTM model to generate the walk encoding.
The sequence of $k$ walk encodings are aggregated to generate a representation for the node $x$.
Temporal Graph Network (TGN) \cite{rossi2020temporal}  generates node embeddings for dynamic graphs that are modeled as a stream of edges. This model consists of several modules.
\begin{itemize}
    \item \textit{Memory}:  At each time $t$, TGN saves a vector $s_i(t)$ for each node $i$ in the memory to represent the node history in a compressed format. The vector $s_i(t)$ is initialized as a zero vector and updated as more edges emerge.
    
    \item  \textit{Message Function}: Every time an edge occurs between two nodes, a message is sent to the nodes participating in the edge. 
    
    \item \textit{Message Aggregator}: A node can be involved in multiple interactions. TGN keeps the last message in the order of time and the mean of other messages from other interactions for each node.
    
    \item \textit{Memory Updater}: Every time an event occurs, the memory of the participating nodes is updated using a learnable memory update function such as LSTM.
    
    \item \textit{Embedding}: The embedding module generates the temporal embeddings for each node at any time $t$ using a learnable function $h$ that updates the representation of each node even if the node was not involved in any interaction until that point. 
    
\end{itemize}
In \cite{wang2020streaming},
a streaming GNN is modeled as $G^1,G^2,..., G^T$ where $G^t= G^{t-1}+ \delta G^t$ and $\delta G^t$ is the changes of a graph between times $t-1$ and $t$. 
The loss of the network at time $t$ is formulated as $L_{new} + L_{existing}$. $L_{new}$ is the loss of parts of the graph influenced by the new changes at time $t$. $L_{existing}$ preserves information from previous time points. In $L_{new}$, the influenced nodes by new changes are replayed in the GNN model. 
In $L_{existing}$, the important nodes from the history are replayed. In addition, the model parameters are approximated such that they do not deviate drastically from the model parameters at the previous time.
Temporal Graph Attention Network (TGAT) \cite{xu2020inductive} 
is a dynamic version of GAT \cite{velivckovic2017graph}. GAT does not consider time ordering between neighbors of a node and is used for static settings. However, TGAT assumes ordering between neighbors of a node based on the time they arrive. The assumption is that a neighbor that occurs more recently is likely to have more influence on a node. In order to add time to the attention mechanism, a time vector is concatenated to the node feature vector. 
Time features that are used in TGAT are obtained based on the concepts from Bochner's Theorem and expressed by mapping the time to $\R^d$ is as follows:

\begin{equation}
    \phi_d(t)=\sqrt{\frac{1}{d}}[cos(w_1t),sin(w_1t),...,cos(w_{d/2}t),sin(w_{d/2}t)]
\end{equation}
which parameters $w_1,...,w_{d/2}$ are learnable parameter. 
Causal Anonymous Walks Neural Net (CAW-N) \cite{wang2020inductive}
generates temporal link embeddings by capturing the motif evolution in dynamic networks using a variant of anonymous walks \cite{ivanov2018anonymous}.
This method predicts the probability of a link in the future and assumes that if motif structures of two nodes $u,v$ interact over time, the probability of a link occurrence between $u,v$ is higher.  Therefore, this model defines set-based anonymization on the temporal random walks to capture the interaction between the motifs of the two nodes over time.
Then, in order to obtain a representation of a link $(u,v)$, all the anonymized walks for nodes $u$ and $v$ are encoded and aggregated using mean or attention mechanism and passed through an MLP to obtain the link probability. 
In Motif-preserving Temporal Shift Network (MTSN) \cite{liu2021motif}, 
the dynamic network is considered as a series of graph snapshots over time 
and two components for generating node embeddings at each time point are introduced. The first component is a Motif Preserving Encoder (MPE) and the second is a Temporal shIft based on Motif preserving Encoder (TIME). MPE component preserves the high-order similarity between nodes at each snapshot. First, it generates node embeddings by running a simplified GCN on the adjacency matrix and a combined motif matrix of the graph separately and adding their outputs.
The combined motif matrix of a graph is obtained by weighted averaging of motif matrices of the input graph that are computed using the Parametrized Graphlet Decomposition (PGD) technique \cite{ahmed2015efficient}. 
TIME component considers the effect of time and is inspired by the Temporal Shift Mechanism in computer vision \cite{lin2019tsm}. It shifts the node embeddings at each snapshot to capture the temporal evolution. In \cite{fu2021sdg,hajiramezanali2019variational,liu2021inductive,yang2020featurenorm}, several other dynamic/temporal GNNs are presented.



\subsection{GNN-based method's real-world applications}
Table \ref{application} presents some of the real-world applications of graph neural nets which are deployed in production in several companies.

\begin{table}[]
\caption{Some of the real-world applications of graph neural nets {\bf deployed in production}}
\label{application}
\begin{tabular}{p{5.2cm}p{9cm}}
\hline
The applied Algorithm & Application \\ \hline
 Standard GNN with MetaGradients \cite{derrow2021eta} & Estimated time of travel (ETA) prediction in Google Maps             \\
 HetMatch \cite{liu2021heterogeneous}& Keyword matching for bid keyword recommendation in sponsered search platform of Alibaba Group  \\
 Category-aware GNN   \cite{qu2020category}        &    Review helpfulness prediction in Taobao       \\ 
 GNN based tag ranking (GraphTR) \cite{liu2020graph}& Video recommendation in WeChat Top Stories\\
 DecGCN\cite{liu2020decoupled}& Online recommendation system in JD.com\\
 DHGAT\cite{niu2020dual}&Search matching in shop search in Taobao\\
 Dynamic Heterogeneous GNN\cite{luo2020dynamic}& Real-time event prediction in DiDi platform\\
 Spatio-temporal graph neural net (ConSTGAT)\cite{fang2020constgat}&  Travel time estimation in Baidu Maps\\
 Heterogeneous Graph Attention Maching Network (HGAMN)\cite{huang2021hgamn}& Retrieving point of interests in different languages in Baidu Maps\\
 PinSage\cite{ying2018graph}&Recommendation system at Pinterest\\
 Gemini\cite{xu2020gemini}& Online recommendation at DidiChuXing \\
 M2GRL\cite{wang2020m2grl}&Recommender system at Taobao\\
 \hline
\end{tabular}
\end{table}

\subsection{The limitation of GNNs and the proposed solutions}
\label{limitation}
GNNs have several known limitations such as expressive power, oversmoothing and scalability. In this section, we summarize the major articles that address these issues.


\subsubsection{Expressive power}
The expressive power of a model refers to the model's ability to distinguish between different graphs. In other words, the model can map different graphs to different embeddings and similar graphs to similar embeddings. The expressive power of common GNNs such as GCN \cite{kipf2016semi} and GraphSAGE \cite{hamilton2017inductive} is bounded by the WL test and they fail to distinguish certain non-isomorphic substructures. A more powerful GNN-based embedding method called Graph Isomorphism Network (GIN) is proposed in \cite{xu2018powerful}. GIN employs the sum operator instead of the mean or max operators to aggregate the neighbors of a node using the following formula: 
\begin{equation}
    h_v^{(k)}= \text{MLP}^{(k)}((1+\epsilon^{(k)})h_v^{(k-1)}+\sum_{u \in N(v)}h_u^{(k-1)})
\end{equation}
where $\epsilon$ can be a fixed or learnable parameter. GIN uses the sum aggregator as its expressive power is higher than the mean or max operators. For instance, assume we have two nodes $v_1$ and $v_2$. If $v_1$ has two equal neighbors and $v_2$ has three equal neighbors. The mean aggregator generates the same embedding for two nodes. The same applies to the max aggregator. However, the sum operator distinguishes different graph structures and generates different embeddings. It is also theoretically proven that GIN's expressive power is equal to the WL test.
Identity-aware Graph Neural Networks (ID-GNN) \cite{you2021identity} proposes a coloring mechanism as a solution for increasing the expressive power of GNNs. This model has two variants. The first variant has two components: inductive identity coloring and heterogeneous message passing. For computing the embedding for each node $v$, a k-hop ego networks of node $v$ is extracted and the center node of the ego network is colored. Then, for each node in the ego network of node $v$, the embedding is computed using a different message passing component for each node based on color. This paper proposes a fast variant of ID-GNN that augments the node features instead of coloring nodes by injecting identity information such as cycle counts for cycles that start and end in the node $v$.
Nested Graph Neural Net (NGNN) \cite{zhang2021nested} suggests to encode a rooted subgraph instead of rooted subtree in common GNNs to generate node representations. It argues that rooted subtrees have limited expressiveness to represent non-tree graphs.  In \cite{li2020distance,zhang2018link}, distance-based features are added to each node to increase the expressive power of GNNs, where distance vectors for each node are computed with respect to each center node. Table \ref{expressive} summarizes the major studies related to increasing the expressive power of GNNs.  

\begin{table}[]
\caption{A summary of major solutions proposed to increase the expressive power of GNNs}
\label{expressive}
\begin{tabular}{llp{9.5cm}}
\hline
Authors & Algorithm & Brief summary of the solution \\ \hline
Xu et al. \cite{xu2018powerful}&      GIN     & Aggregating neighbors using the sum operator \\
Murphy et al \cite{murphy2019relational}& RP-GNN& Adding a unique node label\\
Morris et al. \cite{morris2019weisfeiler}&   k-GNN        & Performing message passing between subgraph structures instead of the node level   \\
Maron et al. \cite{maron2019provably}&    PPGN      & Considering higher-order message passing               \\
Chen et al. \cite{chen2019equivalence}&Ring-GNN& Using a ring of matrices in addition and multiplication\\
Azizian et al. \cite{azizian2020expressive}&    FGNN       & Augmenting the model with matrix multiplication               \\
Li et al. \cite{li2020distance}&  DEGNN         & Adding an extra node feature based on distance encoding               \\
Balcilar et al. \cite{balcilar2021breaking}&  GNNML         & Designing the convolution in spectral domain and masking it with a large receptive field\\
Barcelo et al. \cite{barcelo2021graph}& -          & Adding local graph parameter to GNNs              \\
Sato et al. \cite{sato2021random}& rGIN & Adding random features to GIN \cite{xu2018powerful}               \\
Papp et al. \cite{papp2021dropgnn}&  DropGNN         & Dropping some of the nodes randomly              \\
Wang et al. \cite{wang2021equivariant}&  PEG& Using separate channels to update the node and positional features               \\
Wijesinghe et al. \cite{wijesinghe2021new}&   GraphSNN        & Injecting structural characteristics in the message passing\\
Zhang et al. \cite{zhang2021nested}& NGNN &Encoding a rooted subgraph for each node instead of a rooted subtree       \\
You et al. \cite{you2021identity}& ID-GNN & Inductively injecting node identities either using coloring mechanism or an augmented node feature       \\
Dasoulas et al. \cite{dasoulas2021coloring}&CLIP& Using colors to distinguish similar node attributes\\
Huang et al. \cite{huang2022going}&PG-GNN&Capturing correlation between neighboring nodes using a permutation-aware aggregation\\
Wijesinghe et al. \cite{Wijesinghe2022new}&GraphSNN&Designing a local isomorphism hierarchy for nodes neighborhood subgraphs\\
        \hline
\end{tabular}
\end{table}

\subsubsection{Oversmoothing} A critical known issue with GNNs is their depth limitation \cite{li2018deeper,oono2019graph}. GNN methods aggregate information from one-hop neighbors of a node in the first layer. The second layer reaches the two-hop neighbors of a node and stacking additional layers goes forward in the neighbors of a node similarly. After passing through multiple layers, the generated nodes vectors will be \textit{oversmoothed} because the local information for each node is lost. Graph Random Neural Network (GRAND) \cite{feng2020graph} proposes a new framework to address the oversmoothing problem. This method augments the feature matrix of the input graph using DropNode, which randomly drops features of some of the nodes, similar to the DropEdge mechanism \cite{rong2019dropedge}. This augmentation helps in making nodes less sensitive to their neighbor nodes. Then, it aggregates neighbors of a node up to $K$-hop away using mixed-order propagation, which lowers the risk of oversmoothing. The mixed-order propagation formula is  $X = \sum_{k=0}^K \frac{1}{K+1}\hat{A}^k \bar{X}$, where $\bar{X}$ is the augmented feature matrix. The model is trained using consistency regularization \cite{berthelot2019mixmatch} also to reduce the overfitting issue in the semi-supervised setting in the case of scare labels. 
In \cite{chen2020simple}, it has been theoretically proved that adding two simple techniques to GCN at each layer can overcome oversmoothing. First, constructing a connection to input features to ensure at least a fraction of node features reach the final node representation. Second, adding the identity matrix to the weight matrix to enforce at least the same performance as a shallow GCN. The formulation of this deep GCN is as follows:
\begin{equation}
    H^{(l+1)} = \sigma(((1-\alpha_l)\hat{A}H^{(l)}+\alpha_lH^{(0)})((1-\beta_l)I_n+\beta_lW^{(l)}))
\end{equation}
where $\alpha_l$ and $\beta_l$ are the hyperparameters. $H^{(0)}$ are the initial node representations which are the node features. $I_n$ is the identity matrix. 
Table \ref{oversmooth} shows the formulation of some of the major methods that are introduced for alleviating the oversmoothing problem in GNNs.

\begin{table}[]
\caption{A summary of major solutions proposed to alleviate oversmoothing of GNNs. $Z$ is the node prediction label, $H$ is the node representation,  $K_l$ is the convolution of the $l$-th layer, $s$ is a projection vector, $G_o$ is the discrete gradient operator on the graph, $\hat{A}_{drop}$ is the symmetric normalized adjacency matrix with certain number of edges dropped, INFLATION(.,e) = Normal(Power(Softmax(.),e)), $f_l$ is the number of features at layer $l$, $S^l$ is the clustering assignment matrix, $C$ is number of groups, $\mu_i,\sigma_i$ are the mean and standard deviation of group $i$. $\gamma_i,\beta_i, \lambda$ are hyper parameters. TPSD is a total pairwise squared distance measure, $X_{drop}$ is the perturbed feature matrix, $p$ is an step size, $G_v$ is an extracted subgraph for a node $v$. $\Phi_l$ is a learnable parameter, $\tilde{L}$ is symmetric normalized Laplacian matrix. $psi$ is a learnable function, X is the feature matrix. $T$ is time step.}
\label{oversmooth}
\renewcommand{\arraystretch}{1.5}
\begin{tabular}{lp{11.3cm}}
\hline
 Algorithm & Formula \\ \hline
GCN \cite{kipf2016semi}&  $H^{(l+1)}= \sigma(\hat{A}H^lW^l)$\\
APPNP\cite{klicpera2018predict}&$H=\sigma(\hat{A}H^lW^l)$, $Z^{l+1}=(1-\alpha)\hat{A}Z^l+\alpha H$\\
JKNet-Concat \cite{xu2018representation}&$h_v=\text{concat}(h_v^1,...,h_v^L)W$\\
GCN-PN \cite{zhao2019pairnorm}& $H^{(l+1)}=\text{TPSD}(\sigma(\hat{A}H^lW^l))$ \\
DropEdge \cite{rong2019dropedge}&$H^{(l+1)}=\sigma(\hat{A}_{drop}H^lW^l)$\\
SGC \cite{wu2019simplifying}&$H=\hat{A}^LXW$\\
DGN-GNN\cite{zhou2020towards}&$H^{(l+1)}=S^{(l+1)}H^{(l+1)}+ \lambda \sum_{i=1}^C \gamma_i((H_i^{(l+1)}-\mu_i)/\sigma_i)+\beta_i$\\
DAGNN\cite{liu2020towards}& $H=\text{concat}(mlp(X), \hat{A}^1mlp(X),...,\hat{A}^kmlp(X)))$, $Z = \text{softmax}(\sigma(Hs)H)$ \\
GRAND \cite{feng2020graph}&$Z = mlp(1/(L+1)\sum_{i=0}^{L}\hat{A}^iX_{drop})$\\
GCNII \cite{chen2020simple}& $H^{(l+1)} = \sigma(((1-\alpha_l)\hat{A}H^{(l)}+\alpha_lH^{(0)})((1-\beta_l)I_n+\beta_lW^{(l)}))$\\
GDC \cite{hasanzadeh2020bayesian}& $H^{(l+1)}=\sigma(\sum_{i=1}^{f_l}\hat{A}_{drop}[:,i]H^l[:,i]W^l[i,:])$\\
PDE-GCN \cite{eliasof2021pde}& $h^{(l+1)}=h^l-pG^TK_l^T\sigma(K_lG_oh^l)$\\
GRAND-PDE \cite{chamberlain2021grand}& $H=\psi(X(0)+\int_0^T \frac{\partial X(t)}{\partial t}dt)$\\
SHADOW-SAGE \cite{zeng2021decoupling}&$h^{(l+1)}_v=\sigma(W^l. \text{concat}(h^l_v, \text{aggregate} (h_u^l, \forall u \in G_v))$\\
GCN+inflation \cite{he2022inflation}&$H^{(l+1)}= \text{INFLATION}^{(l+1)}( \sigma(\hat{A}H^lW^l),e)$\\
AdaGNN \cite{dong2021adagnn}& $H^{(l+1)}=H^{(l)}-\tilde{L}H^{(l)}\Phi_l$\\
\hline
\end{tabular}
\end{table}

\subsubsection{Scalability}
  Another bottleneck of GNNs is scalability. In GNNs, the representations of a node's neighbors are aggregated to generate the node embeddings. Specifically, for a GNN with $L$ layers, the neighbor aggregation computed by the matrix multiplication $AH^l$ has the time complexity $O(LmF)$, where $m$ is the number of neighbors and $F$ is the hidden dimension of model. The number of neighbors can be large in very large graphs, which lowers GNNs training speed and increases the memory consumption. In \cite{hamilton2017inductive,chen2018stochastic,yao2021blocking}, this problem is alleviated by sampling a subset of node neighbors. In \cite{chen2018fastgcn}, neighbors of a node are sampled at each layer independently. 
  Cluster-GCN \cite{chiang2019cluster} reduces the memory problem of GCN by sampling a subgraph for each batch using clustering techniques and applying a graph convolution filter on the nodes in the subgraph. Some methods such as SGC \cite{wu2019simplifying} remove the non-linear activation function to reduce the training time. 
  RevGNN \cite{li2021training} is based on the reversible connections \cite{gomez2017reversible} and lowers the memory consumption of GNNs with respect to the number of layers. In this model, the feature matrix is divided into several groups, which are then input into the model to generate a group of outputs. The advantage of this model is that only the output of the last input is saved in memory for backpropagation. In \cite{chen2021unified}, a unified GNN sparsification (UGS) framework is proposed that jointly simplifies the graph and the model to lower the GNN's inference time. The loss function of UGS is:
  \begin{equation}
      L_{UGS}=L(\{m_g \odot A,X\},m_\theta\odot\Theta)+\gamma_1|m_g|+\gamma_2|m_{\theta|}
  \end{equation}
  where $m_g,m_{\theta}$ are masks for the unimportant connections in the graph and weights in GNNs. $L$ is the cross-entropy loss and $\gamma_1,\gamma_2$ are hyperparameters to control the regularization of $m_g,m_{\theta}$. After the training, these two masks prune the adjacency matrix and model parameters. Several other papers that studied the scalability of GNNs are \cite{ding2021vq,chen2020scalable,jia2020redundancy,bojchevski2020scaling,huang2021scaling,Feyetal2021,yoon2021performance,peng2021graphangel,huang2018adaptive}. Table \ref{scalability} provides the complexity of GCN and some of the proposed approaches for lowering its complexity.

\begin{table}[]
\caption{Time and memory complexities of GCN and some of the proposed scalable GNN models. $L$ is the number of layers, $F$ is the hidden dimension of the model, $n$ is the number of nodes, $m$ is the number of neighbors per node, $b$ is the batch size, $d$ is the average degree of the graph, $s_n$ is the number of sampled nodes per node,$s_l$ is the number of sampled nodes per layer, $\delta$ is the ratio of blocked nodes, $\Tilde{s}_n=s_n \times (1-\delta)$, $m_g$ is the number of remaining edges, $m_\theta$ is the number of remaining connections in the model, $k$ is number of codewords .
}
\label{scalability}
\begin{tabular}{lp{4cm}p{5.2cm}p{2cm}}
\hline
Method & Solution & Time complexity& Memory complexity \\ \hline
GCN \cite{kipf2016semi} &   -       & $O(LmF+LnF^2)$    & $O(LnF)$        \\
GraphSAGE \cite{hamilton2017inductive} &   Neighbor sampling&   $O(ns_n^LF+ns_n^{L-1}F^2)$  &   $O(s_n^LbF)$    \\
SGC \cite{wu2019simplifying}    & Linear model & $O(nF^2)$    & $O(bF)$             \\
ClusterGCN \cite{chiang2019cluster}  & Graph Sampling &$O(LmF+LnF^2)$ &     $O(LbF)$         \\
FastGCN \cite{chen2018fastgcn}  &Layer sampling &  $O(Lns_lF+LnF^2)$    &  $O(Ls_lbF)$            \\
LADIES \cite{zou2019layer}  &Layer sampling &  $O(Lns_lF+LnF^2)$    &  $O(Ls_lbF)$   \\
GraphSAINT \cite{zeng2019graphsaint} & Graph sampling& $O(LbdF+LnF^2)$     &    $O(LbF)$   \\
VR-GCN \cite{chen2018stochastic}&Graph sampling& $O(LmF+LnF^2+s_n^LnF^2)$&$O(LnF)$\\
GBP \cite{chen2020scalable}&Linear model &    $O(LnF^2)$    & $O(bF)$            \\
RevGNN \cite{li2021training} & Reversible connections &  $O(LmF+LnF^2)$    &   $O(nF)$           \\
VQ-GNN \cite{ding2021vq} & Vector quantization & $O(LbdF+LnF^2+LnkF)$     &$O(LbF+LkF)$              \\
BNS \cite{yao2021blocking} &Neighbor sampling& $O(\Tilde{s}_n^{L-1}.(s_nbF+(\delta/(1-\delta)+1).bF^2))$     &  $O(\Tilde{s}_n^{L-1}s_nbF)$            \\
GLT \cite{chen2021unified} & Graph Sampling and model pruning&$O(Lm_gF+Lm_{\theta}nF^2)$& $O(LnF)$\\
     \hline
\end{tabular}
\end{table}

\subsubsection{Capturing long-range dependencies in graphs} GNNs struggle to capture long-range dependencies between nodes in the graph. The reason is that broadening the GNNs receptive field by increasing their depth encounters the oversmoothing problem in node representations. To solve this problem, in \cite{wu2021representing} a transformer module is combined with the standard GNN to capture the long-range relationships. Efficient infinite-depth graph neural net (EIGNN) \cite{liu2021eignn} has implicit infinite layers which can capture very long dependencies. The output of EIGNN before the softmax layer is a limit of an infinite sequence of convolutions:
\begin{equation}
    f(X,F,B) = B(\lim_{H \rightarrow \infty} Z^{(H)})
\end{equation}
where $Z^{(H)} =\gamma g(F)Z^{(H-1)}S+X$ is the output of the $H$-th layer. $F,B$ are learnable weight parameters, S is the normalized adjacency matrix of the input graph and $\gamma \in (0,1]$. The $g(F)$ function is defined as $\frac{1}{|F^TF|+\epsilon_F}F^TF$ which is constrained to be less than one. As a result, the infinite sequence of convolutions is convergent. A closed-form solution is derived for EIGNN instead of using iterative solvers. In \cite{lukovnikov2021improving}, the depth-wise and breadth-wise propagations are considered to capture long-range dependencies in graphs. The breadth-wise propagation is between the representation of a node with its neighbors' representations from the previous layer, which is a form of horizontal skip-connections. This model leverages a residual message function to consider edge features and alleviate the breadth-wise gradient diminishing in the model's backpropagation.

\subsubsection{Catastrophic Forgetting} Catastrophic forgetting means that the neural network model may forget previously learned knowledge when trained on a new task. In \cite{liu2021overcoming}, a topology-aware weight preserving module (TWP) is proposed to alleviate this issue in GNNs. This module measures the importance of the GNN's parameters after learning each task. Then, it enforces the model to remember the old parameters when learning a new task by penalizing changes to the important parameters with respect to the old tasks. In \cite{zhou2021overcoming}, an experience replay based model (ER-GNN) selects some nodes from previous tasks and replays them when learning new tasks. The replayed nodes are those whose features are the closest to the mean of features in each class and have the maximum coverage and influence in model training.

\subsubsection{Homophily assumption}
GNNs are based on the homophily \cite{mcpherson2001birds} assumption which means that nodes that are connected are similar and have the same class labels. However, this is not true in networks that nodes with opposite characteristics connect to each other. In \cite{zhu2020beyond}, three design principles from previous methods are combined to make the new model suitable for both homophily and heterophily settings. 
\begin{enumerate}
    \item Separating node embeddings and neighbor embeddings because mixing the node and neighbor information results in similar embeddings among a neighborhood. 
    \item Considering higher order Neighborhoods to capture more relevant information from more neighbors. 
    \item Combining the intermediate representations to increase the range of neighbors and information considered in node representations. 
\end{enumerate}
Similarly, it has been shown in \cite{suresh2021breaking} that 
the disassortativity of many real-world graphs can lead to the low performance of GNN models on these graphs. 
Therefore, this paper proposes to generate a computation graph from the original graph 
and then run the GNN on the computation graph. 
The computation graph is a multi-relational graph with different types of edges between two nodes based on different levels of neighborhood similarities such as nodes degrees and neighboring nodes degrees.
In \cite{zhu2021graph}, a class compatibility matrix is added into GNNs to improve the performance of GNNs in graphs that heterophily exist. 
This framework first estimates a prior belief of every node's class label based on the node features. 
Then, using a compatibility matrix $H$, the prior beliefs of nodes are propagated in their neighborhood. Each element $H_{ij}$ in the matrix $H$ is the empirical probability that nodes with class label $i$ connects to nodes with class label $j$ in the dataset. The compatibility matrix which can be learned in this model enables it to go beyond the homophily assumption.
Table \ref{homophil} summarizes some of the major papers that studied the homophily assumption in GNNs.

\begin{table}[]
\caption{A summary of major solutions proposed to make GNNs suitable for both hemophily and heterophily}
\label{homophil}
\begin{tabular}{llp{10.2cm}}
\hline
Author&Algorithm & Brief summary of the solution \\ \hline
Zhu et al. \cite{zhu2020beyond}&H2GCN&Combining three designs principles: 1) separation of ego and neighbor sampling, 2) higher order neighbors, 3) combining intermediate representations.\\
Chien et al. \cite{chien2020adaptive}&GPR-GNN&Propagating node hidden features using generalized pagerank methods\\
Suresh et al. \cite{suresh2021breaking}&WRGNN&Generating a computation graph based on nodes' structural equivalences \\
Yang et al. \cite{yang2021diverse}&DMP&Setting an attribute propagation weight for each edge\\
Zhu et al. \cite{zhu2021graph}&CPGNN& Adding a compatibility matrix\\
Jin et al. \cite{jin2021universal}& U-GCN& Extracting three types of node embeddings from 1-hop, 2-hop and k-nearest neighbors \\
Liu et al. \cite{liu2021non}&NLGNN&Employing an attention-guided sorting of neighbor nodes\\
Yang et al.\cite{yang2022graph}&GPNN& Leveraging a pointer network to rank neighbor nodes\\
Wang et al. \cite{wang2022powerful}&HOG-GCN&Incorporating a learnable homophily degree matrix into a GCN\\
Fang et al. \cite{fang2022polarized}&Polar-GNN&Using dissimilarities between nodes in the aggregation by introducing attitude polarization \\
Du et al. \cite{du2022gbk}&GBK-GNN&Utilizing two kernels to capture the homophily and heterophily and selecting the appropriate one for each node pair\\
Li et al. \cite{li2022finding}&GloGNN&Finding global homophily for nodes showing heterophily by learning a coefficient matrix\\
\hline
\end{tabular}
\end{table}

\subsubsection{Neglecting substructures} It has been shown in \cite{chen2020can} that the expressive power of message passing GNNs in detecting subgraphs of three or more nodes is limited. Therefore, a Local relation pooling (LPR) model is proposed based on egonets. An egonet centered at a node is a subgraph consisting of nodes within a certain distance from the node in the graph. The LPR's idea is that a pattern in a graph can be found in the egonet of some node. Therefore, it generates a node representation by aggregating transformed egonets centered at the node. Graph Structural Kernel Network (GSKN) \cite{long2021theoretically} 
accounts for graph substructures such as cliques or motifs by leveraging anonymous walk-based graph kernels (AWGK). Graph kernels are similarity measures for pairwise graph comparison. This method defines an anonymous walk kernel and a random walk kernel to capture structural information in the graph. These kernels are defined based on the definition of $l$-walk kernels, which compute the similarity between two graphs by comparing all length $l$ walks between every node in two graphs. Then, the GSKN formulation is derived using the kernel mapping of these two kernels. In message passing simplicial networks (MPSN) \cite{bodnar2021weisfeiler}, the message passing is performed on simplicial complexes \cite{nanda21} which are a form of subgraphs consisting of several simplices. For instance, 0-simplex is a node, 1-simplex is an edge and 2-simplex is a triangle. The representation of each simplex is computed by aggregating four types of messages received from its adjacent simplices which are present in the graph, including boundary, co-boundary, lower and upper adjacencies. For example, an edge's boundary simplices are its vertices and the co-boundary simplices of a node are its connected edges. Finally, the global embedding for a simplicial complex is computed using a readout function on the representation of its simplices. In
\cite{pei2019geom,bouritsas2022improving,long2020graph,alsentzer2020subgraph,thiede2021autobahn}, several other GNNs with the focus on the subgraphs are presented. Table \ref{substructure} demonstrates the formulas for some of these methods. 

\begin{table}[]
\caption{The formulas for the major solutions proposed to help GNNs capture substructures. $i$ is a neighborhood , $r$ is a relationship, $\tau(z_v,z_u)$ is a function that defines a relationship from node $v$ to node $u$ in a latent space, $x_v^V$ are the combined structural features of node $v$, $e_{u,v}$ are the edge $(u,v)$ features, $h_p$ is the embedding of a simplex $p$, $m_B(p)$,$m_C(p)$,$m_{\downarrow}(p)$,$m_{\uparrow}(p)$ are the aggregation of messages from the boundary, co-boundary, lower and upper adjacent simplices of the simplex $p$, $R_v$ is a row in a node-topic matrix representing the probabilities of a node $v$ in a graph belonging to the graph's structural topics, $S_{n_{v,t}}^{k-BFS}$ is the set of permutation of subset of nodes in egonet of node $v$ of depth $t$ compatible with k-truncated breadth first search, $\alpha_{\pi}$ is a learnable normalization factor for $\pi$, $f$ can be an MLP layer, $B_{v,t}^{[ego]}$ is the tensor representation of the egonet of node $v$, $\phi$ is an anonymous walk, $R(\phi)$ is the concatenation of the attributes of $\phi$, $Z=[R(\phi_i)]_i$, $\Phi_l$ is the set of anonymous walks of length $l$, $h_{x,c}$ is the representation of a subgraph $c$, $\gamma$ is a learnable similarity measure, $A_x$ is a subgraph at channel $x$, $a_x$ is the representation of $A_x$.}
\label{substructure}
\renewcommand{\arraystretch}{1.5}
\begin{tabular}{p{3cm}p{11.2cm}}
\hline
Algorithm & Formula \\ \hline
Geom-GCN \cite{pei2019geom}&$h_v^{l+1}=\sigma(W_l.\text{aggregate}_{i,r}((e_{(i,r)}^{v,l+1},(i,r)))), e_{(i,r)}^{v,l+1}=\text{aggregate}(\{h_u^l|u \in N_i(v), \tau(z_v,z_u)=r\})$\\
GSN-v \cite{bouritsas2022improving}&$h_v^{l+1}=\sigma(h_v^l, m_v^{l+1}), m_v^{l+1} = \text{aggregate}(\{h_v^l, h_u^l, x_v^V, x_u^V, e_{u,v}\}) $\\
MPSN \cite{bodnar2021weisfeiler}&$h_p^{l+1}=\text{aggregate}(h_p^{l},m_B^l(p),m_C^l(p),m_{\downarrow}^l(p),m_{\uparrow}^l(p))$\\
GraphSTONE \cite{long2020graph} &$h_v^{l+1}=\text{aggregate}(\{\frac{R_v^TR_u}{\sum_uR_v^TR_u}h_u^l, u \in N(v)\})$\\
DeepLPR \cite{chen2020can}&$H_v^{l+1}=\frac{1}{|S_{n_v,t}^{k-BFS}|}\sum_{\pi \in S_{n_{v,t}}^{k-BFS}}\alpha^l_{\pi} \odot f^l(\pi * B_{v,t}^{[ego]}(H^l))$\\
GSKN \cite{long2021theoretically}&$h_v=\sum_{\phi \in \Phi^l(G,v)}\sigma(Z^TZ)^{-1/2}\sigma(Z^TR(\phi))$\\
SUBGNN \cite{alsentzer2020subgraph}&$h_{x,c}^{l+1}=\sigma(W.[g_{x,c}^l;h_{x,c}^l]), g_{x,c}^l=\text{aggregate}(\{\gamma(c,A_x).a_x, \forall A_x\})$\\
\hline
\end{tabular}
\end{table}

\subsubsection{Over-squashing} 
One of the bottlenecks of GNNs is over-squashing when capturing long-range dependencies. The over-squashing is different from the oversmoothing because the oversmoothing occurs when the graph learning task only needs short-range dependencies and stacking more layers in GNN makes the node embeddings indistinguishable. However, when long-range dependencies are required and more layers are added to GNNs,  the information from distant neighbors is compressed in a fixed-length vector, resulting in over-squashing and low performance of GNNs \cite{alon2020bottleneck}. A simple solution for this problem is presented in \cite{alon2020bottleneck} where a fully-adjacent layer is added as the last layer to a GNN model. In this layer, every two nodes are connected which helps consider nodes beyond the nodes' local neighbors in their representation. In \cite{topping2021understanding}, it is proved that negatively curved edges are the cause of over-squashing. A negative curvature occurs when an edge becomes a bridge between two sides of the graph where removing the edge disconnects them. Therefore, a curvature based graph rewiring model is proposed to solve the over-squashing issue. This rewiring approach works by adding extra edges to support the most negatively curved edges and removing the most positively curved edges. Furthermore, the graph edit distance between the original graph and the modified one is bounded to ensure a local graph modification.

\subsubsection{Design space for GNNs} Designing effective architectures for GNNs in different tasks and datasets requires manual labor and domain knowledge. Therefore, several methods were proposed to automatically design suitable architecture for the given datasets and downstream tasks. Graph neural architecture search method (GraphNAS) \cite{gao2020graph} allows the automated architecture design using reinforcement learning. In GraphNAS, an RNN generates a GNN architecture that will be trained and validated on the training and validation sets. The validation result is the reward of the RNN. The RNN samples the design of each layer from a search space of different operators such as neighbor sampling, message computation, message aggregation and readout operators. For example, a sample GNN layer is:
\begin{equation}
[\text{first-order, gat, sum, 4, 8, relu}]    
\end{equation}
where each element in the list corresponds to neighbor sampling, message computation, message aggregation operators, number of heads, number of hidden layers and the activation function. In \cite{you2020design}, a general design space, task space and evaluation method for GNNs are proposed to identify the best GNN architecture for the given task quickly. The design space consists of intra-layer design, inter-layer design and learning configuration. The intra-layer design in each layer of GNN consists of batch normalization, dropout, activation function and aggregation function. The inter-layer design between GNN layers has four dimensions: layer connectivity, pre-process layers, message passing layers and post-processing layers. The training configuration concerns the batch size, learning rate, optimizer and training epochs. Then, a task similarity metric is introduced that can identify similar tasks. Finally, it develops a controlled random search evaluation to quickly find the best GNN design for the given task among many model-task combinations. In \cite{wang2021autogel,huan2021search}, other methods related to the architecture design of GNNs are introduced.

\section{Open Issues and Ongoing Research}
Graph representation learning has been a very active field of study. Although numerous techniques were developed, there are still significant issues and challenges that need to be addressed. We will discuss some of them here.


\textbf{Proposing solutions for limitations of GNNs.} As discussed in section \ref{limitation}, GNNs have several limitations such as expressive power, scalability and over-squashing. Different solutions have been proposed for these issues. However, some of them, such as over-squashing, were newly discovered and solutions to them are still preliminary, which have potential for further research. In addition, the proposed solutions for the limitations of GNNs were all for static GNNs and not much work has been done on solving the problems of dynamic and spatial-temporal GNNs. Furthermore, there might be new issues in dynamic settings that are not discovered yet and are worth studying. 
Finally, by applying GNNs to different real-world problems and datasets, new drawbacks of GNNs may be discovered, which will be interesting to study and solve. 


\textbf{Studying the theory side of GNNs.}
GNNs have been very successful in different applications, and researchers have tried to understand the theoretical aspects of GNNs success. However, the theoretical analyses of GNNs models in terms of optimization properties and generalization across graph sizes are less understood. In \cite{xu2021optimization}, the first steps in understanding the global convergence of gradient descent in GNNs are studied. In addition, in \cite{yehudai2021local}, it is shown that GNNs that are trained on some graph distributions cannot generalize to larger unseen graphs. Therefore, future work is needed to understand these topics fully.

\textbf{Defining new GNNs by employing differential equations.} 
Partial differential equations (PDE) are used to model physical phenomena \cite{brandstetter2021message}. Recently, PDEs have been used to model the information propagation in graphs and design new GNN models \cite{chamberlain2021grand,eliasof2021pde}. One of the advantages of these new GNNs is that they address the over-smoothing limitations of GNNs. Therefore, it is an interesting future work to investigate what other GNN models can be proposed using differential equations and if these models are more powerful than other GNNs.

\textbf{Including domain-knowledge into GNNs.} The domain knowledge is the information about a specific problem that may not be available for a machine learning model. This knowledge can be included in the models by changing the input, loss function, and model architecture \cite{dash2022review}. For instance, we can enrich the input data by considering additional relationships or constraints on existing relationships among the entities in the data.
A few works incorporate domain knowledge into GNNs, such as \cite{dash2022inclusion,dash2021incorporating}, but this area is still new and has the potential for further research.



\textbf{Limited training data labels.} 
Training neural networks with limited training data has always been challenging, which is also true for training the graph neural nets. 
Solutions such as self-supervised learning, data augmentation, and contrastive learning, have been proposed to solve the label scarcity problem and are employed in graph neural nets. However, there are still opportunities for further studying these types of learning in GNNs in both static and dynamic settings.

\textbf{Transferring advances in deep learning models to GNNs.} Graph neural nets use deep learning techniques. Any advances and new models that are proposed in deep learning can be adapted in graph neural nets. Specially, computer vision and natural language processing are very active areas and the new models that are developed for images, videos and texts can be studied in graphs as well. For instance, ideas from video representation learning might be useful in dynamic graph learning because the concept of objects moving in a video's frames is similar to nodes changing over time.





\textbf{More applications.}
GNNs achieved great success in application in many domains such as social networks, financial networks, and protein structures. Some of the newer applications of GNNs are in electrical power grid monitoring \cite{ringsquandl2021power}, drug overprescription prediction \cite{zhang2021rxnet}, and paper publication prediction \cite{guan2021vpalg}. Therefore, it is interesting to investigate the effectiveness of GNNs in other applications.


\section{Conclusion}
In this survey, we reviewed the node/graph embedding methods, which can be divided into traditional and GNN-based methods. Different from previous surveys on graph representation learning, we provide the literature review in both traditional and GNN-based graph embedding methods for both static and dynamic graphs and include the recent papers published until the time of this submission. Traditional static methods generate node embedding vectors for static graphs and are categorized into factorization based, random walk based and non-GNN based deep learning models. 
Traditional dynamic methods capture the evolving patterns in the history of nodes and are based on aggregations, random walks, non-GNN deep learning and temporal point processes. In addition to traditional methods, GNN based methods have achieved huge success in generating node representations. GNNs are deep learning models which generate a node representation by aggregating the node's neighbors embeddings and often use information from the particular graph mining task of interest in learning the node representations. We reviewed the general framework of GNNs and their categories, including static GNNs, spatial-temporal GNNs and dynamic GNNs and their real-world applications. Furthermore, we summarize nine limitations of GNNs and the proposed solutions to these limitations. These limitations are expressive power, over-smoothing, scalability, over-squashing, capturing long-range dependencies, design space, neglecting substructures, homophily assumptions, and catastrophic forgetting. Previous surveys does not provide such a summary. Finally, we discussed some of the open issues in the node representation learning field that are interesting future research directions. These future directions include proposing solutions for limitations of GNNs in both static and dynamic setting, learning GNNs with limited training data labels in both static and dynamic setting, transferring advances in deep learning to GNNs and further studying GNN's theory side.
As this field is growing fast, we hope that this up-to-date survey can provide valuable additional information for researchers working in this area.



\bibliographystyle{ACM-Reference-Format}
\bibliography{sample-base2}


\end{document}